\documentclass{article} 
\usepackage{iclr2027_conference,times}
\iclrfinalcopy
\usepackage{graphicx}

\usepackage{amsmath,amsfonts,bm}

\def\eqref#1{equation~\ref{#1}}

\def\1{\bm{1}}

\def\vzero{{\bm{0}}}

\def\vk{{\bm{k}}}

\def\vm{{\bm{m}}}

\def\vq{{\bm{q}}}

\def\vs{{\bm{s}}}

\def\vv{{\bm{v}}}

\def\vy{{\bm{y}}}

\def\evalpha{{\alpha}}

\def\mF{{\bm{F}}}

\def\mS{{\bm{S}}}

\DeclareMathAlphabet{\mathsfit}{\encodingdefault}{\sfdefault}{m}{sl}
\SetMathAlphabet{\mathsfit}{bold}{\encodingdefault}{\sfdefault}{bx}{n}

\def\sI{{\mathbb{I}}}

\def\emA{{A}}

\newcommand{\Ls}{\mathcal{L}}
\newcommand{\R}{\mathbb{R}}

\newcommand{\softmax}{\mathrm{softmax}}
\newcommand{\sigmoid}{\sigma}

\usepackage{hyperref}
\usepackage{url}
\usepackage{graphicx}
\usepackage{booktabs}   
\usepackage{multirow}   
\usepackage{pifont}     
\usepackage{titletoc}
\usepackage{placeins}
\usepackage{caption}
\usepackage{xcolor}
\usepackage{soul}
\usepackage{tabularx}
\usepackage{subcaption}
\usepackage{booktabs}
\usepackage{pifont}
\usepackage{booktabs}
\usepackage{tabularx}
\usepackage{arydshln}   
\usepackage{booktabs, multirow, makecell, pifont, xcolor, colortbl, graphicx}
\newcommand{\cmark}{\textcolor{green!55!black}{\ding{51}}}
\newcommand{\xmark}{\textcolor{red!70!black}{\ding{55}}}
\newcommand{\na}{\textcolor{gray}{--}}
\definecolor{tsarow}{RGB}{232,244,234}

\sethlcolor{yellow}   
\title{TSA: Temporal Slot Activation for Persistent Object-Centric Video Representation}

\author{%
\parbox[t]{\dimexpr\textwidth-2\tabcolsep\relax}{\raggedright
Duc~Nguyen$^{1*}$, Sieu~Tran$^{1*}$, Hao~Vo$^{1}$, Khoa~Vo$^{1}$,
Duy~Minh~Ho~Nguyen$^{2}$, Nghi~D.~Q.~Bui$^{3}$, Anh~Nguyen$^{4}$,
Long~Mai$^{5}$, Ngan~Le$^{1}$} \\
$^{1}$University of Arkansas, USA \\
$^{2}$Max Planck Research School for Intelligent Systems \\
$^{3}$Google Research, Google \quad
$^{4}$University of Liverpool, UK \quad
$^{5}$Adobe Research \\
$^{*}$Equal contribution
}

\begin{document}

\maketitle
\lhead{Preprint} 
\begin{abstract}
Unsupervised video object-centric learning aims to decompose dynamic scenes into temporally persistent entity representations. Existing recurrent video slot-attention methods propagate a fixed set of slots across frames, but typically assume \emph{unconditional slot propagation}: every slot is updated and decoded at every frame, regardless of whether its corresponding object is visible. We show that this design violates a basic lifecycle requirement for persistent slots: when an object is absent or fully occluded, its slot should preserve its previous state and avoid explaining unrelated visible content. Instead, unconditional propagation creates \emph{update-induced state drift}, where current-frame evidence overwrites the absent object's representation, and \emph{decoder-induced reconstruction interference}, where the inactive slot remains coupled to reconstruction through decoder attention. We propose \textbf{Temporal Slot Activation (TSA)}, a lightweight mechanism that learns a per-slot, per-frame activation score $\evalpha_{k,t} \in (0,1)$ without visibility supervision. TSA uses this activation as a shared latent control variable for slot lifecycle modeling. When a slot is inactive, TSA anchors its state to the previous slot through activation-gated updating and suppresses its decoder participation through an activation-dependent additive bias on attention logits before softmax normalization. This jointly reduces state drift and reconstruction-driven interference. To improve decisions under partial occlusion and gradual reappearance, TSA further conditions activation prediction on a per-slot temporal memory produced by a Temporal Context Encoder. We evaluate TSA on MOVi-C, MOVi-E, YouTube-VIS HQ, and the occlusion-heavy OVIS benchmarks, using both standard metrics (FG-ARI, mBO) and tracking-based metrics (IDF1, HOTA). TSA improves object decomposition and temporal identity preservation, with large gains on long, heavily occluded videos. Source code will be released upon publication.
\end{abstract}

\section{Introduction}
\label{sec:intro}
Humans perceive visual scenes as collections of persistent objects that remain identifiable through motion, occlusion, and reappearance~\citep{spelke2007core,kahneman1992reviewing}. Object-centric learning (OCL) aims to recover such structure without supervision by decomposing visual inputs into entity-level representations~\citep{greff2019multi,eslami2016attend,burgess2019monet,engelcke2019genesis,lin2020space}. Slot Attention (SA)~\citep{locatello2020object} has become a standard formulation for OCL and a foundation for compositional reasoning and prediction tasks~\citep{wu2022slotformer,wu2023slotdiffusion,kakogeorgiou2024spot,seitzer2022bridging}. Extending SA from images to videos introduces a central requirement, \emph{temporal consistency}: a slot should preserve the same object identity across time, including under partial or full occlusion. To this end, Video Slot Attention (VSA) methods propagate slot states forward and update them using the current frame~\citep{kipf2021conditional, elsayed2022savi++, zadaianchuk2023object, manasyan2025temporally, zhao2026predicting}.

\begin{figure}[t]
    \centering
    \includegraphics[width=\linewidth]{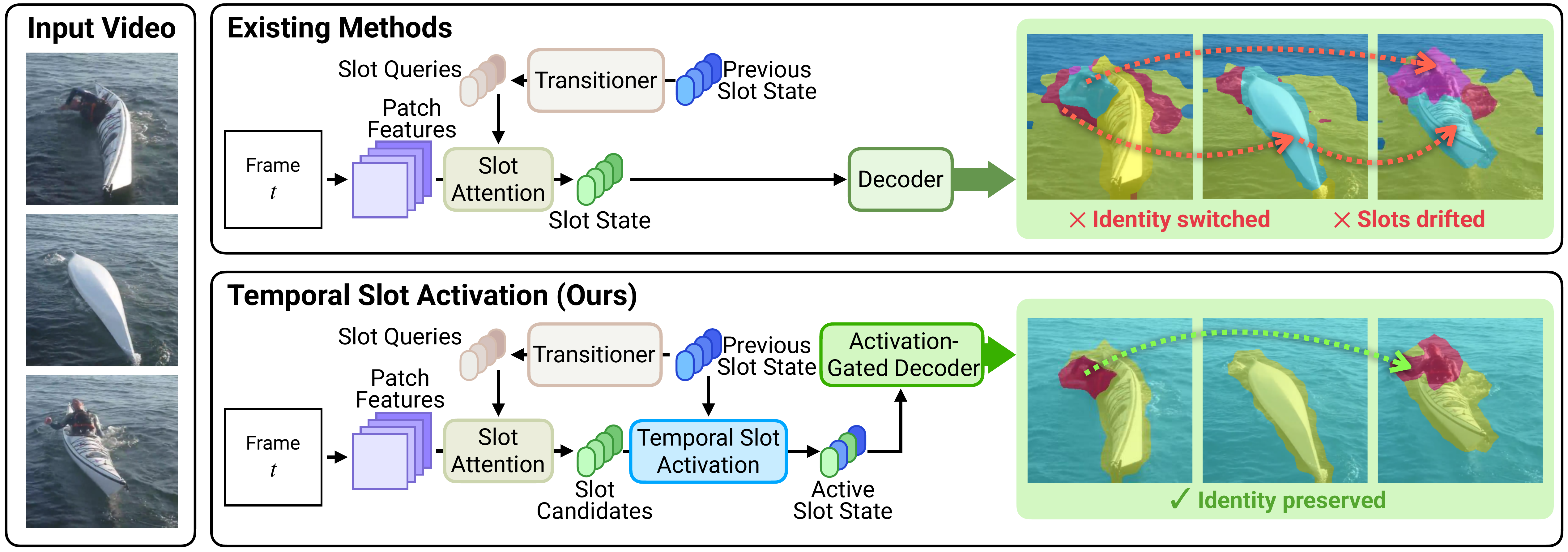}
        \caption{\textbf{Unconditional slot propagation vs.\ TSA under occlusion.} \emph{Top:} Without activation gating, the kayaker's slot drifts toward the occluding hull and triggers an identity switch. \emph{Bottom:} TSA deactivates the absent slot via $\alpha_{k,t}$, preserving its state for consistent reacquisition.}
    \label{fig:teaser}
\vspace{-1em}
\end{figure}

\noindent
Despite their effectiveness, these methods share a common structural assumption that we call \emph{unconditional slot propagation}: every slot is updated and decoded at every frame, regardless of whether the corresponding object is currently visible. This assumption conflates object persistence with visual presence, leading to \emph{state drift} under occlusion. Fig.~\ref{fig:teaser} illustrates this failure: when a kayaker becomes fully occluded by a capsized kayak, the competition in Slot Attention forces every slot to align with some visible content, so the kayaker's slot is reassigned to the occluding kayak hull. This overwrites the previously stored object representation with unrelated features, and the slot gradually loses the identity of the object it was tracking. When the object reappears, the corrupted slot state can no longer function as a meaningful query for reacquisition, and the object is instead captured by another slot, resulting in an \emph{identity switch}. In this sense, representation drift is the underlying mechanism, while identity switch is its observable consequence. This drift is jointly driven by two coupled mechanisms: \emph{unconditional state update} and \emph{unconditional decoder participation} (Sec.~\ref{sec:problem}).

\noindent
We address this problem by introducing \textbf{Temporal Slot Activation~(TSA)}, a lightweight mechanism that assigns each slot $k$ at frame $t$ a learned \emph{activation score} $\alpha_{k,t}\in(0,1)$, trained without visibility supervision. The activation score serves as a shared latent control variable governing the slot lifecycle. For slot-state evolution, TSA performs an \emph{activation-gated state update}: active slots ($\alpha_{k,t}\to 1$) focus on the current SA candidate, whereas inactive slots ($\alpha_{k,t}\to 0$) remain anchored to their previous states, preventing occlusion-induced overwriting. For decoding, TSA performs \emph{activation-gated decoder participation} by applying an additive log-bias on cross-attention logits before softmax, suppressing inactive slots during decoder competition.
Through this dual gating, TSA enforces consistent inactive-but-persistent behavior, with an inactive slot simultaneously protected from current-frame updates and prevented from explaining unrelated visible content.

\noindent
We evaluate TSA on the standard video OCL benchmarks MOVi-C, MOVi-E~\citep{greff2022kubric}, and YouTube-VIS HQ~\citep{ke2022video}. To further examine identity persistence in more challenging settings, we additionally evaluate on \textbf{OVIS}~\citep{qi2022occluded}, which features crowded scenes, long videos, severe occlusions, and objects that frequently disappear and reappear. On all benchmarks, we report the conventional grouping metrics FG-ARI and mBO together with the identity-sensitive tracking metrics HOTA~\citep{luiten2021hota} and IDF1~\citep{ristani2016performance}.

\noindent
Our contributions are: \textbf{(i)} We identify \emph{unconditional slot propagation} as a \textbf{fundamental limitation} of recurrent VSA methods. We provide a formal analysis showing how it causes \emph{representation drift} and \emph{identity switch} through two coupled mechanisms: \emph{update-induced state drift} and \emph{decoder-induced reconstruction interference} (Sec.~\ref{sec:problem}). \textbf{(ii)} We propose \textbf{Temporal Slot Activation~(TSA)}, a lightweight mechanism equipping each slot with a learned per-frame \emph{activation score}~$\alpha_{k,t}$ that jointly controls slot-state evolution and decoder participation, enabling inactive-but-persistent slot behavior (Sec.~\ref{sec:method}). \textbf{(iii)} We extend the evaluation protocol with an occlusion-centric benchmark and identity-sensitive tracking metrics, making long-term slot persistence explicitly measurable (Sec.~\ref{sec:experiments}).

\section{Related Work}
\label{sec:related}

\noindent \textbf{Object-centric learning (OCL) and Slot Attention (SA).}
OCL aims to represent a visual scene as a set of discrete object-level entities without supervision.
Early approaches achieved this through sequential attention or iterative inference
mechanisms that extract objects from an image~\citep{eslami2016attend,burgess2019monet,greff2019multi}.
SA~\citep{locatello2020object} later introduced a scalable alternative based on competitive cross-attention, where a fixed set of slots compete to explain the scene, and has since become the dominant paradigm in OCL. Subsequent work has primarily focused on improving \emph{per-frame decomposition quality}.
These improvements come from stronger pretrained visual features~\citep{oquab2023dinov2,seitzer2022bridging}, more expressive generative decoders~\citep{wu2023slotdiffusion,jiang2023object}, and flexible slot parameterizations~\citep{fan2024adaptive,liu2026metaslot}.
These advances are developed for the single-image setting and form the basis for extensions to video.

\noindent \textbf{Slot utilization.}
A related direction asks how many slots a scene needs. SOLV~\citep{aydemir2023self} merges slots with high cosine similarity into one. AdaSlot~\citep{fan2024adaptive} predicts which slots to keep and masks the rest in the decoder. MetaSlot~\citep{liu2026metaslot} maps slots to a prototype codebook and removes duplicates. All three judge redundancy at inference from the current frame alone. Others target the same problem through training instead, by regularizing redundant slots or increasing the slot budget over the course of training~\citep{seong2026vicious,moon2026reconstruction}.

\noindent \textbf{Video Slot Attention (VSA).}
Extending beyond per-frame decomposition, the video setting requires each slot to consistently represent the same object across frames. SAVi~\citep{kipf2021conditional} addressed this by propagating slot states over time using a learned transition function, followed by refinement with SA at each frame. Subsequent work improves robustness by incorporating additional cues such as depth~\citep{elsayed2022savi++}, discrete tokens~\citep{singh2022simple}, and stronger pretrained features~\citep{zadaianchuk2023object}. Beyond architectural design, another line of work focuses on improving \emph{temporal consistency through training objectives}. For example, VideoSAUR~\citep{zadaianchuk2023object} introduces temporal feature-similarity
losses, while SlotContrast~\citep{manasyan2025temporally} enforces slot identity consistency via
contrastive learning. Others refine the propagation itself.
RandSF.Q~\citep{zhao2026predicting} improves temporal prediction by conditioning transitions on
sampled slot-feature pairs, and SmoothSA~\citep{zhao2026smoothsa} smooths SA recurrences across frames by
preheating cold-start first-frame queries and using fewer iterations on subsequent frames.

Despite these advances, all prior recurrent VSA methods share the same regime of
\emph{unconditional propagation}: every slot is updated and decoded at every frame.
Existing mitigations of temporal inconsistency act on \emph{how} slots are propagated,
through transition dynamics~\citep{kipf2021conditional,zhao2026predicting},
query initialization and iteration schedules~\citep{zhao2026smoothsa},
or temporal objectives~\citep{zadaianchuk2023object,manasyan2025temporally}, but not on
\emph{whether} each slot should be updated or decoded. Slot-utilization
methods withhold slots, but on a different criterion: they judge redundancy against the
visible content, under which an object hidden behind an occluder is indistinguishable from one
that is not in the scene, so the slot holding its identity is released rather than preserved
for reappearance, as detailed in Appendix~\ref{app:slot_control}. Thus, slot-level object correspondence remains implicit, without a
controlled mechanism for preserving object identity in challenging temporal scenarios such as
occlusion. In contrast, TSA introduces a learned \emph{activation score} that explicitly
determines whether each slot is updated and decoded at each frame while preserving
inactive slots, an axis of control orthogonal to slot propagation and
slot-utilization methods.
\section{Limitations of Unconditional Slot Propagation}
\label{sec:problem}

\noindent \textbf{Recurrent VSA Pipeline.}
Let $\vs_{k,t} \in \R^{d}$ denote the state of slot $k \in \{1,\ldots,K\}$ at frame $t \in \{1,\ldots,T\}$, with slot dimension $d$, and let $\mF_t \in \R^{N \times d_f}$ denote the $N$ features of dimension $d_f$ extracted by a frozen visual encoder at time $t$. Given the previous slot states $\mS_{t-1}=\{\vs_{k,t-1}\}_{k=1}^{K}$, a Temporal Query Transitioner ($T_\phi$) 
predicts a query $\vq_{k,t} = T_\phi(\mS_{t-1}, \mF_t)$ for each slot. Existing VSA methods~\citep{kipf2021conditional,elsayed2022savi++,singh2022simple,zadaianchuk2023object,manasyan2025temporally,zhao2026predicting} typically adopt SA, denoted $U_\theta$, to align each query with current-frame evidence via competitive cross-attention:
\begin{equation}
\vs_{k,t} = \mathrm{SA}(\mF_t, \vq_{k,t}) = U_\theta\big(\mF_t, T_\phi(\mS_{t-1}, \mF_t)\big).
\label{eq:unconditional_update}
\end{equation}
All slots are then passed to the decoder. Let $\vq^{d}_{n} \in \R^{d_k}$ be the decoder query at position $n$, and let $\vk^{d}(\vs_{k,t}) \in \R^{d_k}$, $\vv^{d}(\vs_{k,t})$ be the key and value projected from slot $k$. The decoder attention logits and weights are
\begin{equation}
z_{k,n,t} = \frac{1}{\sqrt{d_k}}\, \vq^{d\top}_{n} \vk^{d}(\vs_{k,t}),
\qquad
\emA^{d}_{k,n,t} = \frac{\exp(z_{k,n,t})}{\sum_{j=1}^{K}\exp(z_{j,n,t})}.
\label{eq:decoder_attention}
\end{equation}
We refer to Eq.~\ref{eq:unconditional_update} as \emph{unconditional state update} and to Eq.~\ref{eq:decoder_attention} as \emph{unconditional decoder participation}, because $\emA^{d}_{k,n,t}>0$ for all $k, n, t$, a strict positivity that causes every slot to enter the softmax competition of the decoder at every frame. Together they form the \emph{unconditional slot propagation} regime. To assess this regime, we adopt the lens of \emph{object persistence}: each slot is expected to represent the identity of one object over time. Let $o_{k,t} \in \{0,1\}$ denote the visibility of the object represented by slot $k$ at frame $t$, with $o_{k,t}=1$ when the object is visible and $o_{k,t}=0$ when it is absent or fully occluded. We say slot $k$ is \emph{active} at frame $t$ if $o_{k,t}=1$ and \emph{inactive} if $o_{k,t}=0$.
An active slot should update using the current-frame evidence, whereas an inactive slot should remain persistent, preserving identity for future reappearance.
Unconditional slot propagation violates this expectation, giving rise to two structurally distinct failure pathways analyzed below.

\noindent \textbf{Failure Pathway I: Update-Induced State Drift.}
\label{subsec:update_drift}
Consider an interval $\sI_{[a,b]}=\{a,\ldots,b\}$ during which the object associated with slot $k$ is absent. Object persistence requires $\vs_{k,t} = \vs_{k,t-1}$ throughout $\sI_{[a,b]}$, since $\mF_t$ may contain information about other visible objects and the background, but no evidence for object $k$. Under \emph{unconditional state updating} $\vs_{k,t} = U_\theta(\mF_t, T_\phi(\mS_{t-1}, \mF_t))$, however, state drift accumulates over the absence interval:
\begin{equation}
\left\| \vs_{k,b} - \vs_{k,a-1} \right\|
\;\leq\;
\sum_{t=a}^{b} \left\| \vs_{k,t} - \vs_{k,t-1} \right\|
\;=\;
\sum_{t=a}^{b} \left\| U_\theta\big(\mF_t, T_\phi(\mS_{t-1}, \mF_t)\big) - \vs_{k,t-1} \right\|.
\label{eq:cumulative_drift}
\end{equation}
Because each update is computed from $\mF_t$ and, through $T_\phi$, from the already-drifted state, successive changes tend to point in the same direction rather than cancel, so even small frame-to-frame changes can lead to substantial deviation from the pre-occlusion identity as the absence duration increases.

\noindent \textbf{Failure Pathway II: Decoder-Induced Reconstruction Interference.}
\label{subsec:decoder_interference}
One might attempt to address update-induced drift by freezing the slot state during absence. However, Eq.~\ref{eq:decoder_attention} shows that $\emA^{d}_{k,n,t} > 0$ for all $k,n,t$, so every slot contributes to the decoded output even when the corresponding object is absent. This creates a training-time reconstruction pathway from the loss to the inactive slots. For a reconstruction loss $\Ls_{\mathrm{recon}}(\hat{\vy}_{t}, \vy_{t})$, where $\hat{\vy}_{t} = \{\hat{\vy}_{n,t}\}_{n=1}^{N}$ and $\vy_{t} = \{\vy_{n,t}\}_{n=1}^{N}$ denote the decoded and target features at frame $t$, the derivative with respect to slot $\vs_{k,t}$ presents: 
\begin{equation}
\frac{\partial \Ls_{\mathrm{recon}}}{\partial \vs_{k,t}}
= \sum_{n} \frac{\partial \Ls_{\mathrm{recon}}}{\partial \hat{\vy}_{n,t}}
\frac{\partial \hat{\vy}_{n,t}}{\partial \vs_{k,t}},
\quad \text{where} \quad
\frac{\partial \hat{\vy}_{n,t}}{\partial \vs_{k,t}}
= \emA^{d}_{k,n,t} \frac{\partial \vv^{d}(\vs_{k,t})}{\partial \vs_{k,t}}
+ \sum_{j=1}^{K} \vv^{d}(\vs_{j,t}) \frac{\partial \emA^{d}_{j,n,t}}{\partial \vs_{k,t}}.
\label{eq:decoder_gradient}
\end{equation}
Since $\emA^{d}_{k,n,t}$ is strictly positive, the inactive slot remains coupled to the reconstruction. Consequently, the decoder can use information from an inactive slot to reduce reconstruction error for unrelated visible content. The training objective thus propagates gradients through inactive slots, optimizing the model parameters in a way that may undermine inactive-but-persistent behavior.

\noindent \textbf{Design Requirement.}
The above analysis shows that the two failure pathways are structurally distinct. A valid solution must jointly regulate two conditions: (i) Should slot $k$ update from the current frame? and (ii) Should slot $k$ participate in reconstructing the current frame? For an inactive slot, the desired behavior is
\begin{equation}
\underbrace{o_{k,t} = 0 \Rightarrow \vs_{k,t} \approx \vs_{k,t-1}}_{\text{(A) Pathway I}}
\qquad
\underbrace{o_{k,t} = 0 \Rightarrow \emA^{d}_{k,n,t} \approx 0 \;\; \forall n}_{\text{(B) Pathway II}}.
\label{eq:close_pathway}
\end{equation}
Condition~(A) prevents update-induced state drift, while condition~(B) removes the inactive slot from decoder competition and suppresses reconstruction-driven interference. This motivates a shared activation variable $\evalpha_{k,t} \in (0,1)$ that jointly controls both pathways via \emph{activation-gated state update} and \emph{activation-gated decoder participation}: $\evalpha_{k,t} \to 0 \Rightarrow \{\vs_{k,t} \approx \vs_{k,t-1} \text{ and } \emA^{d}_{k,n,t} \to 0\}$. Using a single activation variable is important for inactive-but-persistent slot behavior, because if state updating and decoder participation were controlled independently, one pathway could remain active while the other is suppressed.
\section{Temporal Slot Activation}
\label{sec:method}

\begin{figure}[t]
    \centering
    \includegraphics[width=0.85\linewidth]{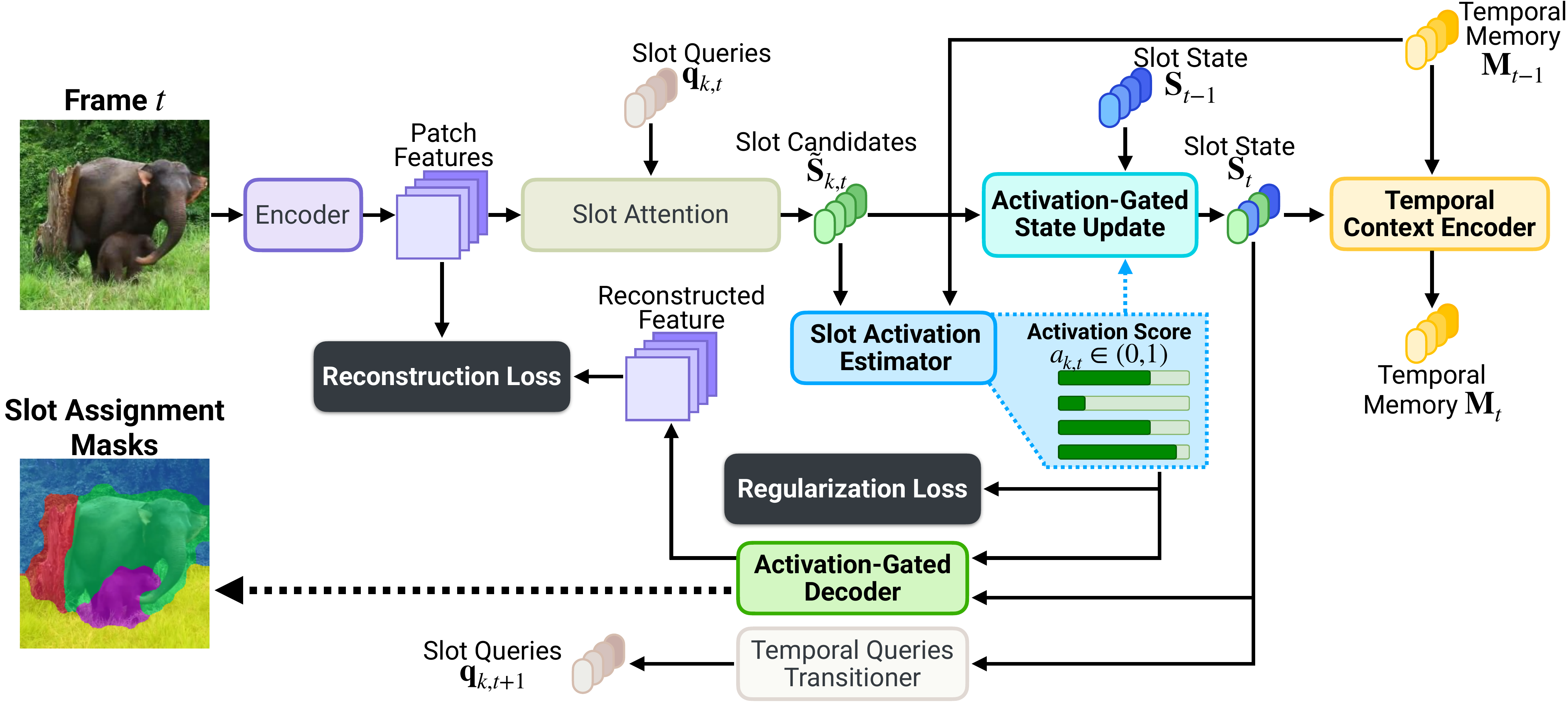}
    \caption{\textbf{Overview of Temporal Slot Activation (TSA).} At each frame $t$, Slot Attention refines slot queries $\vq_{k,t}$ into slot candidates $\tilde{\vs}_{k,t}$, from which the Slot Activation Estimator predicts a learned activation score $\evalpha_{k,t}$. The score jointly controls state updates (Eq.~\ref{eq:state_gate}) and decoder attention (Eq.~\ref{eq:log_bias}), freezing and silencing inactive slots while allowing active ones to track normally.}
    \label{fig:overview}
\vspace{-1em}
\end{figure}

We instantiate the design requirement in Sec.~\ref{sec:problem} with \textbf{Temporal Slot Activation (TSA)}. Each slot $k$ at frame $t$ is equipped with a learned scalar \emph{activation score} $\evalpha_{k,t} \in (0,1)$, trained without visibility supervision. When the slot is \emph{active} ($\evalpha_{k,t} \to 1$), it updates its state and contributes to reconstruction normally. When \emph{inactive} ($\evalpha_{k,t} \to 0$), $\evalpha_{k,t}$ simultaneously \emph{freezes} the slot state (satisfying Eq.~\ref{eq:close_pathway}(A)) and \emph{silences} the slot in the decoder (satisfying Eq.\ref{eq:close_pathway}(B)). Fig.~\ref{fig:overview} illustrates the complete forward pass.

\subsection{Slot activation estimator}
\label{subsec:estimator}
Given the slot query $\vq_{k,t}$ from the transition module $T_\phi$, SA ($U_{\theta}$) refines it using current-frame features $\mF_t$ to produce a candidate state $\tilde{\vs}_{k,t} = U_\theta(\mF_t, \vq_{k,t})$. TSA then predicts the activation score with a shared Slot Activation Estimator $\Phi_{\mathrm{act}}$:
\begin{equation}
\evalpha_{k,t} = \sigmoid\!\left( \Phi_{\mathrm{act}}\big( \tilde{\vs}_{k,t},\, \vm_{k,t-1} \big) \right),
\label{eq:activation}
\end{equation}
where $\vm_{k,t-1} \in \R^{d_h}$ denotes the \emph{temporal memory} of slot $k$, a recurrent summary of its trajectory $[\vs_{k,1}, \ldots, \vs_{k,t-1}]$ produced by a Temporal Context Encoder $\Psi_{\mathrm{tce}}$ (Sec.~\ref{subsec:memory}), capturing the slot's accumulated behavior over time. We condition $\Phi_{\mathrm{act}}$ on the slot candidate $\tilde{\vs}_{k,t}$ rather than the query $\vq_{k,t}$ because the candidate is directly grounded in current-frame evidence. The query is inherited from $\vs_{k,t-1}$ and remains object-like even when the object is absent, making it a weak deactivation signal. In contrast, $\tilde{\vs}_{k,t}$ reflects current-frame evidence directly: when the object is absent, SA fails to align the slot to any region of the object it was tracking, producing a candidate weakly aligned with the slot's own object that serves as a signal for deactivation. The temporal memory $\vm_{k,t-1}$ supplements this with trajectory context, enabling more robust decisions in ambiguous regimes, such as partial occlusion or gradual reappearance, where $\tilde{\vs}_{k,t}$ alone may be misleading.

\subsection{Activation-gated state update}
\label{subsec:state}
To prevent update-induced state drift, TSA replaces state adoption with an activation-gated update:
\begin{equation}
\vs_{k,t} = \evalpha_{k,t}\, \tilde{\vs}_{k,t} + (1 - \evalpha_{k,t})\, \vs_{k,t-1}.
\label{eq:state_gate}
\end{equation}
When the slot is active, $\evalpha_{k,t} \to 1$ and the model incorporates the current candidate. When the slot is inactive, $\evalpha_{k,t} \to 0$ and the update reduces to $\vs_{k,t} \to \vs_{k,t-1}$, satisfying condition~(A) of Eq.~\ref{eq:close_pathway}. The candidate is still computed for activation prediction, since it provides the primary deactivation signal to the estimator $\Phi_{\mathrm{act}}$, but its ability to overwrite the stored slot state is controlled by $\evalpha_{k,t}$.

\subsection{Activation-gated decoder participation}
\label{subsec:decoder}
The activation-gated state update (Sec.~\ref{subsec:state}) alone does not remove an inactive slot from the reconstruction pathway: the slot may still enter the decoder softmax and receive reconstruction-driven gradients. TSA therefore uses the same activation score to modulate decoder attention. Let $z_{k,n,t}$ denote the decoder attention logit between slot $k$ and spatial position $n$. TSA injects the activation score as a pre-softmax additive bias:
\begin{equation}
\emA^{d}_{k,n,t} = \softmax_{k}\!\left(z_{k,n,t} + \log \evalpha_{k,t}\right)
= \frac{\evalpha_{k,t}\exp(z_{k,n,t})}{\sum_{j=1}^{K}\evalpha_{j,t}\exp(z_{j,n,t})}.
\label{eq:log_bias}
\end{equation}
Placing the log-bias inside the softmax embeds activation directly into slot competition: as $\evalpha_{k,t}$ decreases, the contribution of slot $k$ is multiplicatively downweighted before normalization. In the inactive limit ($\evalpha_{k,t} \to 0$), the biased logit $z_{k,n,t} + \log \evalpha_{k,t} \to -\infty$, yielding $\emA^{d}_{k,n,t} \to 0$ for all positions. Thus the same scalar that gates state updates also suppresses decoder participation, aligning decoding with the desired inactive behavior.

This mechanism acts as a learned, continuous analogue of attention masking. Unlike fixed binary masks, TSA uses a dynamic slot-wise gate: intermediate $\evalpha_{k,t}$ softly attenuates uncertain slots, while $\evalpha_{k,t} \to 0$ enforces hard exclusion. Because gating occurs before softmax, it removes both pathways through which inactive slots affect reconstruction: their direct contribution vanishes, and their influence on normalization disappears. Consequently, decoder gating not only preserves inactive slots in the state space but also prevents them from explaining unrelated visible content.

\subsection{Temporal context encoder}
\label{subsec:memory}
Single-frame evidence is often unreliable for activation, especially under occlusion where residual features can produce convincing but incorrect slot candidates. To mitigate this, each slot maintains a \emph{temporal memory} $\vm_{k,t} \in \R^{d_h}$ that summarizes its recent trajectory, providing $\Phi_{\mathrm{act}}$ with historical context to complement the current-frame signal $\tilde{\vs}_{k,t}$. For this memory to be useful upon reappearance, it must be shielded from current-frame evidence during absence; otherwise drift would corrupt the trajectory context needed for reactivation, the memory-level analogue of the state-level drift analyzed in Sec.~\ref{subsec:update_drift}. We compute $\vm_{k,t}$ via a Temporal Context Encoder $\Psi_{\mathrm{tce}}$ conditioned on the \emph{post-gate} slot state $\vs_{k,t}$:
\begin{equation}
\vm_{k,t} = \Psi_{\mathrm{tce}}(\vm_{k,t-1},\, \vs_{k,t}), \qquad \vm_{k,0} = \vzero.
\label{eq:memory}
\end{equation}
The key design choice is conditioning on $\vs_{k,t}$ rather than $\tilde{\vs}_{k,t}$. When the slot is inactive, Eq.~\ref{eq:state_gate} ensures $\vs_{k,t} \approx \vs_{k,t-1}$, so $\Psi_{\mathrm{tce}}$ keeps receiving the preserved pre-absence state rather than current-frame evidence: the activation gate that protects the slot state simultaneously protects the memory from unrelated content, without any additional mechanism. Conditioning on $\tilde{\vs}_{k,t}$ instead would expose the memory to current-frame evidence during inactivity, reintroducing through the memory pathway the same overwrite problem that Eq.~\ref{eq:state_gate} closes at the state level. To prevent unintended interference, $\vm_{k,t}$ is routed exclusively to $\Phi_{\mathrm{act}}$, with no direct connection to the decoder or the transition module $T_\phi$.

\subsection{Training objectives}
\label{subsec:objectives}

TSA is trained with a reconstruction loss $\Ls_{\mathrm{recon}}$~\citep{zadaianchuk2023object} and a slot-consistency contrastive loss $\Ls_{\mathrm{ssc}}$~\citep{manasyan2025temporally}, augmented with an activation regularizer $\Ls_{\mathrm{reg}}$ composed of two complementary terms. Without regularization, $\Ls_{\mathrm{recon}}$ alone admits two degenerate solutions: \emph{full-activation collapse} ($\evalpha_{k,t} \equiv 1$), where TSA reduces to unconditional propagation, and \emph{ambiguous gating} ($\evalpha_{k,t} \approx 0.5$), where neither pathway is decisively controlled. We address both failure modes with a single regularizer,
\begin{equation}
\small
\Ls_{\mathrm{reg}} = \Ls_{\mathrm{usage}} + \beta\, \Ls_{\mathrm{sparse}},
 \text{ where } 
\Ls_{\mathrm{usage}} = \frac{1}{KT}\sum_{k,t} \evalpha_{k,t} \text{ and }
\Ls_{\mathrm{sparse}} = \frac{1}{KT}\sum_{k,t} \evalpha_{k,t}(1-\evalpha_{k,t}),
\label{eq:activation_reg}
\end{equation}
where $\Ls_{\mathrm{usage}}$ penalizes mean activation, creating pressure to deactivate slots that do not improve reconstruction and thereby preventing full-activation collapse; $\Ls_{\mathrm{sparse}}$ penalizes intermediate activation, sharpening decisions toward near-binary behavior and preventing ambiguous gating; and $\beta$ controls the relative weight between them. Full-activation collapse ($\evalpha_{k,t} \equiv 1$) satisfies $\Ls_{\mathrm{sparse}} = 0$ but maximizes $\Ls_{\mathrm{usage}}$, while ambiguous gating ($\evalpha_{k,t} \approx 0.5$) maximizes $\Ls_{\mathrm{sparse}}$ but keeps $\Ls_{\mathrm{usage}}$ at a moderate level. Neither degenerate mode can minimize both simultaneously, so the combined $\Ls_{\mathrm{reg}}$ drives activations toward sparse, near-binary behavior. The full training objective is
\begin{equation}
\Ls = \Ls_{\mathrm{recon}} + \lambda_{\mathrm{ssc}}\, \Ls_{\mathrm{ssc}} + \lambda_{\mathrm{reg}}\, \Ls_{\mathrm{reg}}.
\label{eq:total_loss}
\end{equation}
\section{Experiments}
\label{sec:experiments}

\subsection{Experimental setup}
\label{subsec:setup}

\noindent \textbf{Datasets and metrics.}
Following standard protocols~\citep{zadaianchuk2023object,manasyan2025temporally,zhao2026predicting, zhao2026smoothsa}, we evaluate on MOVi-C, MOVi-E, and YouTube-VIS HQ. We additionally include OVIS to stress-test persistence under severe occlusion and crowded scenes. We report FG-ARI~$\uparrow$ and mBO~$\uparrow$ as standard object-centric metrics, and further include HOTA~$\uparrow$ ~\citep{luiten2021hota} and IDF1~$\uparrow$ ~\citep{ristani2016performance} to directly assess temporal association quality. Dataset details are provided in Appendix~\ref{app:datasets} and full metric definitions in Appendix~\ref{app:eval}.


\noindent \textbf{Implementation details.}
All experiments use a frozen DINOv2 ViT-S/14~\citep{oquab2023dinov2} encoder.
Frames are provided at $256\times256$ and resized to $224\times224$ for the encoder, with slot budgets $K \in \{11, 24, 7, 22\}$ for MOVi-C, MOVi-E, YouTube-VIS HQ, and OVIS respectively.
$\Phi_{\mathrm{act}}$ is a two-layer MLP and $\Psi_{\mathrm{tce}}$ a single-layer GRU.
Full details are in Appendix~\ref{app:impl}, and sensitivity to $\lambda_{\mathrm{reg}}$ and $\beta$ in Appendix~\ref{app:sensitivity}.

\subsection{Main results}

\label{subsec:main}
\begin{table}[t]
\caption{Results on synthetic benchmarks. Mean $\pm$ std over 3 seeds. Best in \textbf{bold}, second best \underline{underlined}. \vspace{-1em}}
\label{tab:main_synthetic}
\begin{center}
\setlength{\tabcolsep}{4pt}
\resizebox{\linewidth}{!}{%
\begin{tabular}{l|cccc|cccc|c|c}
\toprule
\multirow{2}{*}{\textbf{Method}}
  & \multicolumn{4}{|c|}{\textbf{MOVi-C} ($K=11$) \textit{(simple, 24 frames)}}
  & \multicolumn{4}{|c|}{\textbf{MOVi-E} ($K=24$) \textit{(complex, 24 frames)}}
  & \multirow{2}{*}{\textbf{Params (M)$\downarrow$}} & \multirow{2}{*}{\textbf{FPS}$\uparrow$} \\
\cmidrule(lr){2-5}\cmidrule(lr){6-9}
  & FG-ARI$\uparrow$ & mBO$\uparrow$ & HOTA$\uparrow$ & IDF1$\uparrow$
  & FG-ARI$\uparrow$ & mBO$\uparrow$ & HOTA$\uparrow$ & IDF1$\uparrow$
  & \\
\midrule
VideoSAUR
  & $53.3_{\pm2.1}$ & $16.1_{\pm0.4}$ & $17.8_{\pm0.6}$ & $8.1_{\pm1.6}$
  & $34.6_{\pm20.7}$ & $8.3_{\pm4.9}$ & $9.8_{\pm3.9}$ & $3.2_{\pm1.5}$
  & 25.1 & 344 \\
SlotContrast
  & $59.9_{\pm5.3}$ & $27.7_{\pm3.0}$ & $32.1_{\pm3.2}$ & $29.7_{\pm6.9}$
  & $70.6_{\pm3.8}$ & $20.7_{\pm1.4}$ & $22.8_{\pm1.3}$ & $10.7_{\pm3.5}$
  & 31.4 & 295\\
RandSF.Q\textsubscript{tsim}
  & $66.3_{\pm1.7}$ & $28.4_{\pm1.3}$ & $32.8_{\pm1.7}$ & $32.4_{\pm3.2}$
  & $74.0_{\pm1.3}$ & $22.9_{\pm0.9}$ & $25.9_{\pm1.9}$ & $16.3_{\pm4.5}$
  & 34.1 & 411\\
RandSF.Q\textsubscript{ssc}
  & $67.4_{\pm2.1}$ & $29.2_{\pm3.8}$ & \underline{$33.9_{\pm3.9}$} & $\mathbf{33.1_{\pm6.5}}$
  & \underline{$82.1_{\pm3.1}$} & $23.0_{\pm1.2}$ & $25.9_{\pm1.2}$ & $14.9_{\pm1.5}$
  & 34.1 & 411\\
SmoothSA
  & \underline{$69.0_{\pm0.3}$} & $\mathbf{31.7_{\pm0.8}}$ & $26.9_{\pm0.6}$ & $25.4_{\pm1.3}$
  & $73.6_{\pm0.6}$ & $\mathbf{28.6_{\pm0.1}}$ & \underline{$27.5_{\pm0.2}$} & $\mathbf{20.1_{\pm0.4}}$
  & 35.3 & 373 \\
\textbf{TSA (ours)}
  & $\mathbf{75.1_{\pm0.4}}$ & \underline{$29.9_{\pm1.1}$} & $\mathbf{34.9_{\pm1.3}}$ & \underline{$32.7_{\pm1.8}$}
  & $\mathbf{83.6_{\pm0.9}}$ & \underline{$25.1_{\pm0.6}$} & $\mathbf{27.8_{\pm0.7}}$ & \underline{$16.8_{\pm0.9}$}
  & 34.2 & 359\\
\bottomrule
\end{tabular}}
\end{center}
\vspace{-1em}
\end{table}

\begin{table}[t]
\caption{Results on real-world benchmarks. Mean $\pm$ std over 3 seeds. Best in \textbf{bold}, second best \underline{underlined}. \vspace{-1em}}
\label{tab:main_real}
\begin{center}
\setlength{\tabcolsep}{6pt}
\resizebox{\linewidth}{!}{%
\begin{tabular}{l|cccc|cccc}
\toprule
\multirow{2}{*}{\textbf{Method}}
  & \multicolumn{4}{c|}{\textbf{YouTube-VIS HQ} ($K=7$) \textit{(simple, up to 36 frames)}}
  & \multicolumn{4}{c}{\textbf{OVIS} ($K=22$) \textit{(complex, up to 500 frames)}} \\
\cmidrule(lr){2-5}\cmidrule(lr){6-9}
  & FG-ARI$\uparrow$ & mBO$\uparrow$ & HOTA$\uparrow$ & IDF1$\uparrow$
  & FG-ARI$\uparrow$ & mBO$\uparrow$ & HOTA$\uparrow$ & IDF1$\uparrow$ \\
\midrule
VideoSAUR
  & $49.2_{\pm0.5}$ & $29.9_{\pm0.4}$ & $16.9_{\pm0.3}$ & $6.3_{\pm0.1}$
  & $23.4_{\pm0.4}$ & $14.1_{\pm0.2}$ & $5.8_{\pm0.1}$ & $1.4_{\pm0.1}$ \\
SlotContrast
  & $49.4_{\pm1.1}$ & $33.0_{\pm0.2}$ & $18.8_{\pm0.2}$ & $8.7_{\pm0.5}$
  & $24.3_{\pm0.6}$ & $16.1_{\pm0.6}$ & $6.5_{\pm0.4}$ & $1.5_{\pm0.1}$ \\
RandSF.Q\textsubscript{tsim}
  & $60.4_{\pm2.3}$ & \underline{$39.4_{\pm0.3}$} & \underline{$23.8_{\pm0.4}$} & \underline{$19.3_{\pm1.3}$}
  & $22.5_{\pm6.2}$ & $16.2_{\pm3.4}$ & \underline{$8.1_{\pm1.0}$} & \underline{$4.3_{\pm0.2}$} \\
RandSF.Q\textsubscript{ssc}
  & $58.0_{\pm1.0}$ & $37.6_{\pm0.4}$ & $21.6_{\pm0.2}$ & $15.1_{\pm0.6}$
  & $30.4_{\pm0.9}$ & $18.6_{\pm0.7}$ & $7.6_{\pm0.3}$ & $3.0_{\pm0.2}$ \\
SmoothSA
  & \underline{$63.0_{\pm3.4}$} & $38.9_{\pm0.7}$ & $22.7_{\pm0.4}$ & $16.2_{\pm0.5}$
  & \underline{$34.4_{\pm2.4}$} & \underline{$19.7_{\pm0.5}$} & \underline{$8.1_{\pm0.3}$} & $3.9_{\pm0.2}$ \\
\textbf{TSA (ours)}
  & $\mathbf{76.3_{\pm1.2}}$ & $\mathbf{54.2_{\pm1.2}}$ & $\mathbf{44.3_{\pm4.5}}$ & $\mathbf{46.7_{\pm5.9}}$
  & $\mathbf{61.2_{\pm0.2}}$ & $\mathbf{34.0_{\pm1.4}}$ & $\mathbf{36.2_{\pm2.6}}$ & $\mathbf{36.9_{\pm2.5}}$ \\
\bottomrule
\end{tabular}}
\end{center}
\vspace{-1em}
\end{table}


\noindent \textbf{Synthetic benchmarks.}
As shown in Table~\ref{tab:main_synthetic}, TSA achieves the best FG-ARI and HOTA on both
MOVi-C and MOVi-E, the two metrics that most directly reflect object grouping and temporal
association, respectively. On MOVi-C, TSA improves FG-ARI from 69.0 to 75.1 and HOTA
from 33.9 to 34.9 over the strongest competing method on each metric, while ranking second on
mBO (29.9 vs.\ 31.7) and IDF1 (32.7 vs.\ 33.1). The improvement is particularly pronounced in
FG-ARI, indicating that activation-aware slot updating improves foreground object decomposition.
On the more crowded MOVi-E benchmark, TSA further improves FG-ARI from 82.1 to 83.6 and HOTA
from 27.5 to 27.8, again ranking second on mBO and IDF1. Notably, TSA is the only method that
ranks in the top two on all eight metrics across both benchmarks, whereas the baselines exhibit
clear trade-offs: SmoothSA attains the best mBO but falls behind on MOVi-C tracking
(26.9 HOTA, 25.4 IDF1, below even SlotContrast), and RandSF.Q\textsubscript{ssc} leads MOVi-C
IDF1 but drops to 14.9 IDF1 on MOVi-E. These results show that TSA delivers consistently strong
performance across both grouping and tracking under synthetic multi-object dynamics, even though
short clips of rigid objects offer limited opportunity for the inactive-but-persistent regime
that TSA is designed for.

\noindent \textbf{Real-world benchmarks.}
The benefits of TSA become more pronounced on real-world videos (Table~\ref{tab:main_real}), where occlusion, clutter, and long-term dynamics are prevalent. On YouTube-VIS HQ, TSA delivers large gains across both grouping and tracking metrics (e.g.\ $+27.4$ IDF1), indicating improvements not only in per-frame segmentation but also in temporal consistency. This advantage is further amplified on OVIS, a benchmark characterized by heavy occlusion and long trajectories, where TSA substantially outperforms prior methods (e.g.\ HOTA improves from 8.1 to 36.2). These results reinforce a key insight: unconditional slot propagation breaks down in realistic settings, whereas the ability of TSA to deactivate and preserve slots enables more reliable object discovery and identity tracking under complex, occlusion-heavy dynamics.

\subsection{Further Analysis}
\label{subsec:analysis}

\begin{figure}[t]
\centering
\begin{minipage}[t]{0.52\linewidth}
\vspace{0pt}
\captionof{table}{Temporal persistence under varied invisible intervals ($T_\Delta$) on OVIS. Results from a single seed.}
\label{tab:occlusion}
\centering
\setlength{\tabcolsep}{2.5pt}
\resizebox{\linewidth}{!}{%
\begin{tabular}{l|cc|cc|cc|cc}
\toprule
\multirow{2}{*}{\textbf{Method}}
& \multicolumn{2}{c|}{$T_\Delta{=}0$}
& \multicolumn{2}{c|}{$1$--$10$}
& \multicolumn{2}{c|}{$11$--$20$}
& \multicolumn{2}{c}{${>}20$} \\
\cmidrule(lr){2-3}\cmidrule(lr){4-5}\cmidrule(lr){6-7}\cmidrule(lr){8-9}
  & HOTA & IDF1 & HOTA & IDF1 & HOTA & IDF1 & HOTA & IDF1 \\
\midrule
VideoSAUR    & 4.6 & 0.8 & 3.4 & 0.5 & 2.4 & 0.3 & 2.9 & 0.5 \\
SlotContrast & 5.1 & 0.7 & 3.8 & 0.7 & 2.9 & 0.4 & 2.8 & 0.3 \\
RandSF.Q\textsubscript{ssc} & 6.2 & 1.8 & 4.3 & 1.0 & 3.6 & 0.7 & 2.9 & 0.5 \\
SmoothSA & 7.2 & 2.6 & 4.7 & 1.3 & 4.0 & 1.1 & 3.7 & 0.9 \\
\textbf{TSA (ours)}
  & \textbf{36.6} & \textbf{38.9}
  & \textbf{30.3} & \textbf{31.6}
  & \textbf{27.5} & \textbf{29.8}
  & \textbf{23.7} & \textbf{22.0} \\
\bottomrule
\end{tabular}}
\end{minipage}
\hfill
\begin{minipage}[t]{0.45\linewidth}
\vspace{0pt}
\centering
\includegraphics[width=\linewidth]{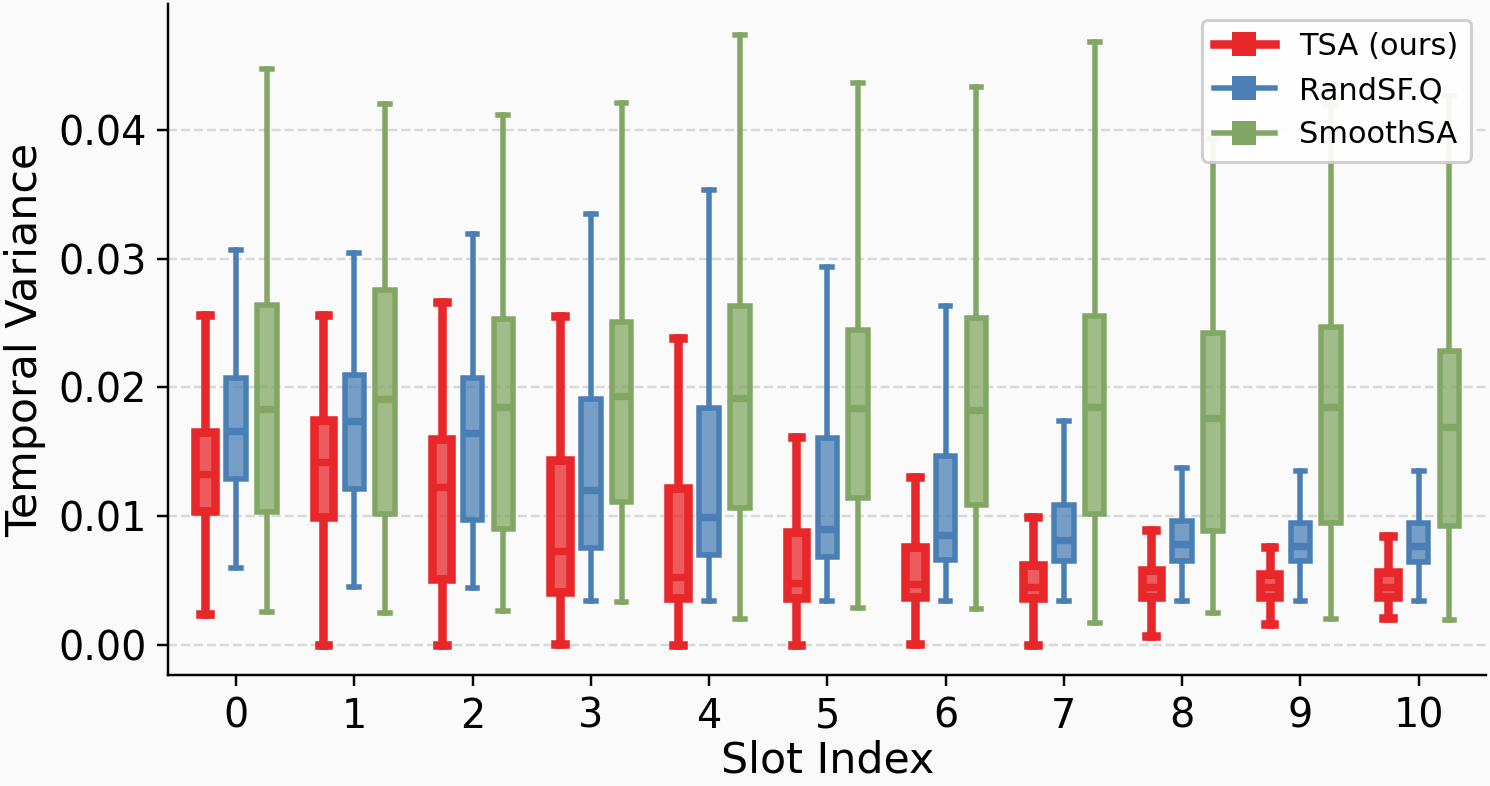}
\vspace{-1em}
\caption{\textbf{Temporal variance per slot.}}
\label{fig:temporal_variance}
\end{minipage}
\vspace{-1em}
\end{figure}

\noindent \textbf{Occlusion duration.}
Table~\ref{tab:occlusion} evaluates persistence under varying lengths of disappearance.
Performance generally degrades as occlusion becomes longer, reflecting the difficulty
of maintaining identity over extended gaps. However, TSA outperforms prior methods by a wide
margin in every regime. Even under the most challenging long-duration occlusions
($T_\Delta > 20$), TSA achieves 23.7 HOTA and 22.0 IDF1, higher than what any prior method
achieves at any occlusion length, including objects that are never occluded ($T_\Delta = 0$).
This highlights the central benefit of TSA: by allowing slots to become inactive while
preserving their internal state, it maintains identity through absence rather than forcing
erroneous updates. As a result, TSA achieves more robust object persistence and
re-identification, especially when objects undergo prolonged occlusion or delayed reappearance.

\noindent \textbf{Representation stability.}
Fig.~\ref{fig:temporal_variance} reports the per-slot frame-to-frame variation of slot states in
MOVi-C. TSA yields consistently lower medians than both RandSF.Q and SmoothSA across all slots, indicating that its slot states evolve more smoothly over time. This confirms
that the activation-gated state update reduces update-induced slot drift by anchoring inactive
slots to their previous states, with lower frame-to-frame variation reflecting more stable
identity preservation. Appendix~\ref{app:drift} provides further evidence: when an object is
occluded and later reappears, the slot of TSA changes less between these two moments than that
of RandSF.Q, on all four datasets.

\subsection{Ablation studies}
\label{subsec:ablation}
We conduct ablation studies on the OVIS benchmark. Additional ablation studies on YouTube-VIS HQ
are provided in Appendix~\ref{app:ablation_ytvis}; further analysis and downstream-task
evaluation in Appendix~\ref{app:downstream}.

\noindent \textbf{A. Effect of activation-gated state update and decoder participation.}
Table~\ref{tab:ablation}~(left) studies the two pathways controlled by $\evalpha_{k,t}$. The baseline (Exp.~\#1), corresponding to unconditional slot propagation, obtains 30.5 FG-ARI, 18.4 mBO, and 7.4 HOTA. Activation-gated decoding alone (Exp.~\#2) leaves all three metrics nearly unchanged (30.0 FG-ARI, 19.1 mBO, 7.9 HOTA), indicating that suppressing inactive slots in the decoder is insufficient when their states are overwritten by current-frame evidence. Activation-gated state update alone (Exp.~\#3) improves FG-ARI to 61.5, mBO to 23.3, and HOTA to 21.2, confirming that gating the state update is essential to prevent occlusion-induced slot drift. The full model (Exp.~\#4) attains comparable FG-ARI while further improving mBO to 34.4 and HOTA to 39.4, supporting our design that state evolution and reconstruction should be \emph{jointly} controlled by a shared activation score.

\noindent \textbf{B. Effect of regularization.}
Table~\ref{tab:ablation}~(middle) shows that reconstruction alone (Exp.~\#1) cannot learn meaningful slot lifecycles, as the model collapses toward unconditional propagation. $\Ls_{\mathrm{sparse}}$ alone (Exp.~\#2) remains at this level, since sharpening activation decisions has little effect when no slot is encouraged to deactivate. $\Ls_{\mathrm{usage}}$ (Exp.~\#3) provides the main gain, raising FG-ARI to 61.0, mBO to 31.7, and HOTA to 26.9 by discouraging unnecessary slot activation and enabling inactive-but-persistent slots. Combining both terms (Exp.~\#4) yields the best performance, improving HOTA further to 39.4 and suggesting complementary roles: $\Ls_{\mathrm{usage}}$ determines \emph{when} slots should be active, while $\Ls_{\mathrm{sparse}}$ makes these decisions more decisive.

\noindent \textbf{C. Effect of temporal memory.}
Table~\ref{tab:ablation}~(right) evaluates temporal context. Removing temporal context (Exp.~\#1) reduces HOTA to 25.4, showing that current-frame evidence alone is unreliable. Conditioning on the previous slot state $\vs_{k,t-1}$ (Exp.~\#2) recovers part of the loss, reaching 60.8 FG-ARI and 33.1 HOTA. Using the accumulated memory $\vm_{k,t-1}$ (Exp.~\#3) performs best, improving HOTA to 39.4. This indicates that activation decisions benefit from longer trajectory context, especially during gradual reappearance and prolonged partial occlusion.

\begin{table}[t]
\caption{Ablation studies on OVIS. \emph{Left:} activation-gated pathways. \emph{Middle:} regularization terms in $\Ls_{\mathrm{reg}}$. \emph{Right:} temporal memory. Results from a single seed. Best per column in \textbf{bold}. \vspace{-1em}}
\label{tab:ablation}
\begin{center}
\setlength{\tabcolsep}{4pt}
\renewcommand{\arraystretch}{1.15}
\begin{minipage}[t]{0.36\linewidth}
\centering
\resizebox{\linewidth}{!}{%
\begin{tabular}{cccccc}
\toprule
\multirow{2}{*}{\textbf{Exp.}} & \textbf{State} & \textbf{Decoder}
& \multicolumn{3}{c}{\textbf{Metrics}} \\
\cmidrule(lr){2-3}\cmidrule(lr){4-6}
 & \textbf{update} & \textbf{participate}
 & FG-ARI$\uparrow$ & mBO$\uparrow$ & HOTA$\uparrow$ \\
\midrule
\#1 & \ding{55} & \ding{55} & 30.5 & 18.4 & 7.4 \\
\#2 & \ding{55} & \ding{51} & 30.0 & 19.1 & 7.9 \\
\#3 & \ding{51} & \ding{55} & \textbf{61.5} & 23.3 & 21.2 \\
\textbf{\#4} & \ding{51} & \ding{51}
 & 61.1 & \textbf{34.4} & \textbf{39.4} \\
\bottomrule
\end{tabular}}
\end{minipage}
\hfill
\begin{minipage}[t]{0.32\linewidth}
\centering
\resizebox{\linewidth}{!}{%
\begin{tabular}{cccccc}
\toprule
\multirow{2}{*}{\textbf{Exp.}} & \multicolumn{2}{c}{\textbf{Loss}}
& \multicolumn{3}{c}{\textbf{Metrics}} \\
\cmidrule(lr){2-3}\cmidrule(lr){4-6}
 & $\Ls_{\mathrm{sparse}}$ & $\Ls_{\mathrm{usage}}$
 & FG-ARI$\uparrow$ & mBO$\uparrow$ & HOTA$\uparrow$ \\
\midrule
\#1 & \ding{55} & \ding{55} & 31.6 & 19.5 & 7.9 \\
\#2 & \ding{51} & \ding{55} & 28.8 & 18.0 & 7.5 \\
\#3 & \ding{55} & \ding{51} & 61.0 & 31.7 & 26.9 \\
\textbf{\#4} & \ding{51} & \ding{51}
 & \textbf{61.1} & \textbf{34.4} & \textbf{39.4} \\
\bottomrule
\end{tabular}}
\end{minipage}
\hfill
\begin{minipage}[t]{0.28\linewidth}
\centering
\resizebox{\linewidth}{!}{%
\begin{tabular}{ccccc}
\toprule
\multirow{2}{*}{\textbf{Exp.}} & \multirow{2}{*}{\shortstack{\textbf{Temporal}\\\textbf{memory}}}
& \multicolumn{3}{c}{\textbf{Metrics}} \\
\cmidrule(lr){3-5}
 & & FG-ARI$\uparrow$ & mBO$\uparrow$ & HOTA$\uparrow$ \\
\midrule
\#1 & \ding{55} & 57.5 & 32.3 & 25.4 \\
\#2 & $\vs_{k,t-1}$ & 60.8 & 29.3 & 33.1 \\
\textbf{\#3} & $\vm_{k,t-1}$
 & \textbf{61.1} & \textbf{34.4} & \textbf{39.4} \\
\bottomrule
\end{tabular}}
\end{minipage}
\end{center}
\end{table}

\subsection{Qualitative results}
\label{subsec:qualitative}

\begin{figure}[t]
    \centering
    \includegraphics[width=\linewidth]{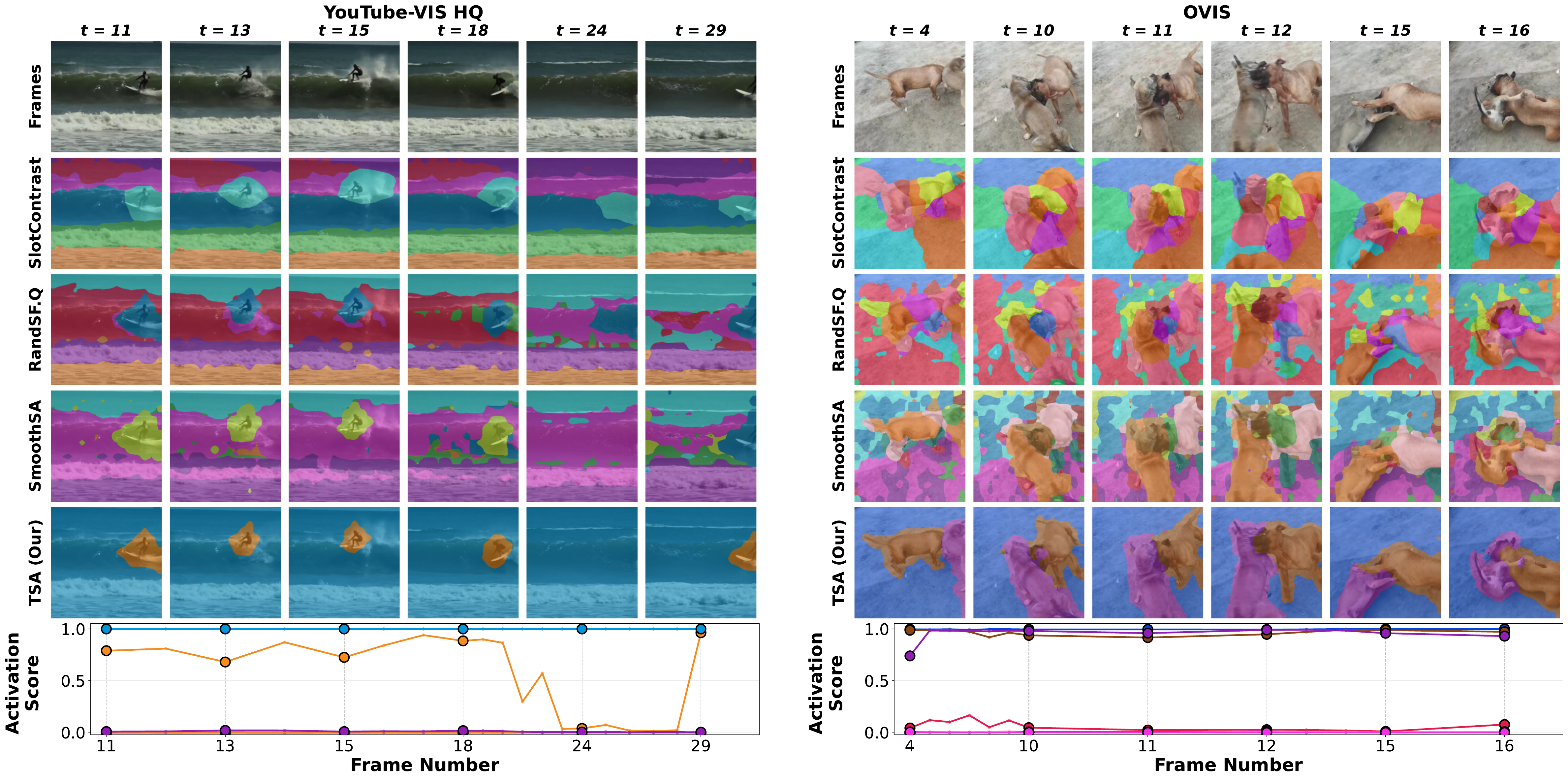}
    \vspace{-1em}
    \caption{\textbf{Qualitative comparison on YouTube-VIS HQ and OVIS.} Colors denote slot identity. The bottom row shows the activation score of each slot predicted by our TSA across frames, with line colors matching the corresponding slot masks.}\vspace{-1em}
    \label{fig:qual_compare}
\end{figure}

Fig.~\ref{fig:qual_compare} presents representative sequences from YouTube-VIS HQ and OVIS. SlotContrast, RandSF.Q, and SmoothSA, all employing unconditional slot propagation, exhibit state drift and identity switches consistent with our analysis in Sec.~\ref{sec:problem}. In contrast, TSA maintains consistent slot identity throughout each sequence. The per-slot activation curves plotted below confirm this behavior: when an object disappears, the activation score of its slot drops to near zero, then rises again upon reappearance, while slots of visible objects remain active. This is consistent with the object lifecycle defined in Sec.~\ref{sec:problem}. Additional qualitative analysis and comparisons are in Appendix~\ref{app:qualitative}.

\section{Conclusion}
\label{sec:conclusion}

We present Temporal Slot Activation (TSA), which shifts unsupervised video object-centric learning from continuous propagation to selective persistence. By identifying \emph{unconditional slot propagation} as a principal cause of state drift and identity switching in recurrent VSA, TSA introduces a simple principle: a slot should update and participate in decoding only when its object is present. A single learned activation score $\evalpha_{k,t}$ jointly gates slot state updates and decoder participation, while a Temporal Context Encoder conditions activation decisions on accumulated trajectory memory, enabling slots to act as stable temporal anchors that preserve object identity through long occlusions and gradual reappearance, without any visibility supervision. TSA remains competitive on the synthetic MOVi-C and MOVi-E benchmarks and delivers substantial gains on the real-world YouTube-VIS HQ and OVIS, with the largest improvements on heavily occluded sequences, offering a lightweight and effective approach to modeling object permanence in unsupervised video learning.

\noindent \textbf{Limitations and future work.}
TSA uses a fixed slot budget $K$; combining activation with scene-adaptive slot allocation remains an open direction. For fair comparison, TSA also builds on a frozen DINOv2 backbone, whose rich features underpin strong performance, though incorporating modalities such as optical flow or depth could further sharpen slot boundaries in cluttered scenes. Finally, while activation gating effectively preserves slot identity through occlusion, gradual appearance changes arising from deformation or scale variation over long sequences remain an open challenge for future work.





\bibliographystyle{iclr2027_conference}
\bibliography{main}

@article{spelke2007core,
  title={Core knowledge},
  author={Spelke, Elizabeth S and Kinzler, Katherine D},
  journal={Developmental Science},
  volume={10},
  number={1},
  pages={89--96},
  year={2007},
  publisher={Wiley}
}

@article{kahneman1992reviewing,
  title={The reviewing of object files: Object-specific integration of information},
  author={Kahneman, Daniel and Treisman, Anne and Gibbs, Brian J},
  journal={Cognitive psychology},
  volume={24},
  number={2},
  pages={175--219},
  year={1992},
  publisher={Elsevier}
}

@inproceedings{greff2019multi,
  title={Multi-object representation learning with iterative variational inference},
  author={Greff, Klaus and Kaufman, Rapha{\"e}l Lopez and Kabra, Rishabh and Watters, Nick and Burgess, Christopher and Zoran, Daniel and Matthey, Loic and Botvinick, Matthew and Lerchner, Alexander},
  booktitle={International conference on machine learning},
  pages={2424--2433},
  year={2019},
  organization={PMLR}
}

@article{eslami2016attend,
  title={Attend, infer, repeat: Fast scene understanding with generative models},
  author={Eslami, SM and Heess, Nicolas and Weber, Theophane and Tassa, Yuval and Szepesvari, David and Hinton, Geoffrey E and others},
  journal={Advances in neural information processing systems},
  volume={29},
  year={2016}
}

@article{burgess2019monet,
  title={Monet: Unsupervised scene decomposition and representation},
  author={Burgess, Christopher P and Matthey, Loic and Watters, Nicholas and Kabra, Rishabh and Higgins, Irina and Botvinick, Matt and Lerchner, Alexander},
  journal={arXiv preprint arXiv:1901.11390},
  year={2019}
}

@article{engelcke2019genesis,
  title={Genesis: Generative scene inference and sampling with object-centric latent representations},
  author={Engelcke, Martin and Kosiorek, Adam R and Jones, Oiwi Parker and Posner, Ingmar},
  journal={arXiv preprint arXiv:1907.13052},
  year={2019}
}

@article{lin2020space,
  title={Space: Unsupervised object-oriented scene representation via spatial attention and decomposition},
  author={Lin, Zhixuan and Wu, Yi-Fu and Peri, Skand Vishwanath and Sun, Weihao and Singh, Gautam and Deng, Fei and Jiang, Jindong and Ahn, Sungjin},
  journal={arXiv preprint arXiv:2001.02407},
  year={2020}
}

@article{locatello2020object,
  title={Object-centric learning with slot attention},
  author={Locatello, Francesco and Weissenborn, Dirk and Unterthiner, Thomas and Mahendran, Aravindh and Heigold, Georg and Uszkoreit, Jakob and Dosovitskiy, Alexey and Kipf, Thomas},
  journal={Advances in neural information processing systems},
  volume={33},
  pages={11525--11538},
  year={2020}
}

@article{wu2022slotformer,
  title={Slotformer: Unsupervised visual dynamics simulation with object-centric models},
  author={Wu, Ziyi and Dvornik, Nikita and Greff, Klaus and Kipf, Thomas and Garg, Animesh},
  journal={arXiv preprint arXiv:2210.05861},
  year={2022}
}

@article{wu2023slotdiffusion,
  title={Slotdiffusion: Object-centric generative modeling with diffusion models},
  author={Wu, Ziyi and Hu, Jingyu and Lu, Wuyue and Gilitschenski, Igor and Garg, Animesh},
  journal={Advances in Neural Information Processing Systems},
  volume={36},
  pages={50932--50958},
  year={2023}
}

@inproceedings{kakogeorgiou2024spot,
  title={Spot: Self-training with patch-order permutation for object-centric learning with autoregressive transformers},
  author={Kakogeorgiou, Ioannis and Gidaris, Spyros and Karantzalos, Konstantinos and Komodakis, Nikos},
  booktitle={Proceedings of the IEEE/CVF Conference on Computer Vision and Pattern Recognition},
  pages={22776--22786},
  year={2024}
}

@article{seitzer2022bridging,
  title={Bridging the gap to real-world object-centric learning},
  author={Seitzer, Maximilian and Horn, Max and Zadaianchuk, Andrii and Zietlow, Dominik and Xiao, Tianjun and Simon-Gabriel, Carl-Johann and He, Tong and Zhang, Zheng and Sch{\"o}lkopf, Bernhard and Brox, Thomas and others},
  journal={arXiv preprint arXiv:2209.14860},
  year={2022}
}

@article{kipf2021conditional,
  title={Conditional object-centric learning from video},
  author={Kipf, Thomas and Elsayed, Gamaleldin F and Mahendran, Aravindh and Stone, Austin and Sabour, Sara and Heigold, Georg and Jonschkowski, Rico and Dosovitskiy, Alexey and Greff, Klaus},
  journal={arXiv preprint arXiv:2111.12594},
  year={2021}
}

@article{elsayed2022savi++,
  title={Savi++: Towards end-to-end object-centric learning from real-world videos},
  author={Elsayed, Gamaleldin and Mahendran, Aravindh and Van Steenkiste, Sjoerd and Greff, Klaus and Mozer, Michael C and Kipf, Thomas},
  journal={Advances in Neural Information Processing Systems},
  volume={35},
  pages={28940--28954},
  year={2022}
}

@article{zadaianchuk2023object,
  title={Object-centric learning for real-world videos by predicting temporal feature similarities},
  author={Zadaianchuk, Andrii and Seitzer, Maximilian and Martius, Georg},
  journal={Advances in neural information processing systems},
  volume={36},
  pages={61514--61545},
  year={2023}
}

@inproceedings{manasyan2025temporally,
  title={Temporally consistent object-centric learning by contrasting slots},
  author={Manasyan, Anna and Seitzer, Maximilian and Radovic, Filip and Martius, Georg and Zadaianchuk, Andrii},
  booktitle={Proceedings of the Computer Vision and Pattern Recognition Conference},
  pages={5401--5411},
  year={2025}
}

@inproceedings{zhao2026predicting,
  title={Predicting video slot attention queries from random slot-feature pairs},
  author={Zhao, Rongzhen and Li, Jian and Kannala, Juho and Pajarinen, Joni},
  booktitle={Proceedings of the AAAI Conference on Artificial Intelligence},
  volume={40},
  number={16},
  pages={13208--13216},
  year={2026}
}

@inproceedings{greff2022kubric,
  title={Kubric: A scalable dataset generator},
  author={Greff, Klaus and Belletti, Francois and Beyer, Lucas and Doersch, Carl and Du, Yilun and Duckworth, Daniel and Fleet, David J and Gnanapragasam, Dan and Golemo, Florian and Herrmann, Charles and others},
  booktitle={Proceedings of the IEEE/CVF conference on computer vision and pattern recognition},
  pages={3749--3761},
  year={2022}
}

@article{qi2022occluded,
  title={Occluded video instance segmentation: A benchmark},
  author={Qi, Jiyang and Gao, Yan and Hu, Yao and Wang, Xinggang and Liu, Xiaoyu and Bai, Xiang and Belongie, Serge and Yuille, Alan and Torr, Philip HS and Bai, Song},
  journal={International Journal of Computer Vision},
  volume={130},
  number={8},
  pages={2022--2039},
  year={2022},
  publisher={Springer}
}

@article{luiten2021hota,
  title={Hota: A higher order metric for evaluating multi-object tracking},
  author={Luiten, Jonathon and Osep, Aljosa and Dendorfer, Patrick and Torr, Philip and Geiger, Andreas and Leal-Taix{\'e}, Laura and Leibe, Bastian},
  journal={International journal of computer vision},
  volume={129},
  number={2},
  pages={548--578},
  year={2021},
  publisher={Springer}
}

@inproceedings{ristani2016performance,
  title={Performance measures and a data set for multi-target, multi-camera tracking},
  author={Ristani, Ergys and Solera, Francesco and Zou, Roger and Cucchiara, Rita and Tomasi, Carlo},
  booktitle={European conference on computer vision},
  pages={17--35},
  year={2016},
  organization={Springer}
}

@article{oquab2023dinov2,
  title={Dinov2: Learning robust visual features without supervision},
  author={Oquab, Maxime and Darcet, Timoth{\'e}e and Moutakanni, Th{\'e}o and Vo, Huy and Szafraniec, Marc and Khalidov, Vasil and Fernandez, Pierre and Haziza, Daniel and Massa, Francisco and El-Nouby, Alaaeldin and others},
  journal={arXiv preprint arXiv:2304.07193},
  year={2023}
}

@article{jiang2023object,
  title={Object-centric slot diffusion},
  author={Jiang, Jindong and Deng, Fei and Singh, Gautam and Ahn, Sungjin},
  journal={arXiv preprint arXiv:2303.10834},
  year={2023}
}

@inproceedings{fan2024adaptive,
  title={Adaptive slot attention: Object discovery with dynamic slot number},
  author={Fan, Ke and Bai, Zechen and Xiao, Tianjun and He, Tong and Horn, Max and Fu, Yanwei and Locatello, Francesco and Zhang, Zheng},
  booktitle={Proceedings of the IEEE/CVF Conference on Computer Vision and Pattern Recognition},
  pages={23062--23071},
  year={2024}
}

@inproceedings{
liu2026metaslot,
title={MetaSlot: Break Through the Fixed Number of Slots in Object-Centric Learning},
author={Hongjia Liu and Rongzhen Zhao and Haohan Chen and Joni Pajarinen},
booktitle={The Thirty-ninth Annual Conference on Neural Information Processing Systems},
year={2025},
}

@article{singh2022simple,
  title={Simple unsupervised object-centric learning for complex and naturalistic videos},
  author={Singh, Gautam and Wu, Yi-Fu and Ahn, Sungjin},
  journal={Advances in neural information processing systems},
  volume={35},
  pages={18181--18196},
  year={2022}
}

@article{aydemir2023self,
  title={Self-supervised object-centric learning for videos},
  author={Aydemir, G{\"o}rkay and Xie, Weidi and Guney, Fatma},
  journal={Advances in Neural Information Processing Systems},
  volume={36},
  pages={32879--32899},
  year={2023}
}

@article{kingma2014adam,
  title={Adam: A method for stochastic optimization},
  author={Kingma, Diederik P and Ba, Jimmy},
  journal={arXiv preprint arXiv:1412.6980},
  year={2014}
}

@inproceedings{ke2022video,
  title={Video mask transfiner for high-quality video instance segmentation},
  author={Ke, Lei and Ding, Henghui and Danelljan, Martin and Tai, Yu-Wing and Tang, Chi-Keung and Yu, Fisher},
  booktitle={European Conference on Computer Vision},
  pages={731--747},
  year={2022},
  organization={Springer}
}

@article{seong2026vicious,
  title={From vicious to virtuous cycles: Synergistic representation learning for unsupervised video object-centric learning},
  author={Seong, Hyun Seok and Moon, WonJun and Heo, Jae-Pil},
  journal={arXiv preprint arXiv:2602.03390},
  year={2026}
}

@article{moon2026reconstruction,
  title={Reconstruction-Guided Slot Curriculum: Addressing Object Over-Fragmentation in Video Object-Centric Learning},
  author={Moon, WonJun and Seong, Hyun Seok and Heo, Jae-Pil},
  journal={arXiv preprint arXiv:2603.22758},
  year={2026}
}

@article{zhao2026smoothsa,
  title={Smoothing slot attention iterations and recurrences},
  author={Zhao, Rongzhen and Yang, Wenyan and Kannala, Juho and Pajarinen, Joni},
  journal={arXiv preprint arXiv:2508.05417},
  year={2025}
}

@misc{luiten2020trackeval,
  author =       {Luiten, Jonathon and Hoffhues, Arne},
  title =        {TrackEval},
  howpublished = {\url{https://github.com/JonathonLuiten/TrackEval}},
  year =         {2020}
}

\clearpage
\appendix

\begin{center}
{\Large \textsc{Appendix}}
\end{center}

\startcontents[appendix]
\printcontents[appendix]{}{1}{}

\clearpage
\appendix

\section{Conceptual comparison of slot-control mechanisms}
\label{app:slot_control}

Sec.~\ref{sec:intro} characterizes recurrent VSA methods by unconditional slot propagation. A separate line of work does assess slot utility and remove slots judged unnecessary, so we situate TSA among all mechanisms that decide whether a slot participates in a frame. Table~\ref{tab:conceptual} classifies them by whether a per-slot, per-frame participation decision is made and the evidence it is conditioned on, which of the two failure pathways of Sec.~\ref{sec:problem} it gates, and the behavior of a slot whose object is fully occluded.

Occlusion requires separating object persistence from slot participation: the slot should stop explaining the current evidence, yet remain available as a query when the object returns. In propagation methods (SAVi~\citep{kipf2021conditional}, VideoSAUR~\citep{zadaianchuk2023object}, SlotContrast~\citep{manasyan2025temporally}, RandSF.Q~\citep{zhao2026predicting}, SmoothSA~\citep{zhao2026smoothsa}) the question never arises, since participation is fixed by construction and every slot competes for the frame whether or not its object is there, leaving both pathways open. Merging and pruning methods do raise it, but resolve it from evidence that occlusion makes ambiguous. SOLV~\citep{aydemir2023self} judges a slot by its similarity to other slots, so a slot that has drifted onto the occluder is merged into the slot already explaining it. AdaSlot~\citep{fan2024adaptive} learns which slots to keep for the current frame, and MetaSlot~\citep{liu2026metaslot} prunes slots that collapse onto the same learned prototype; both judge a slot by what the current frame supports, a criterion an absent object and a hidden one satisfy equally poorly. The verdict is also final, with the slot absorbed into another, or withheld from the decoder with no record of the object instance it held and no criterion for restoring it.

The two failures are opposite and leave a lifecycle gap: a slot is either kept fully active or released, with nothing in between. TSA fills this gap by conditioning the decision on the slot candidate, obtained by querying the current frame with the propagated slot state, together with the slot's temporal memory, rather than on frame content alone, and by making the decision reversible. TSA may also deactivate slots that are simply not needed; what distinguishes it is what deactivation means. A slot with $\alpha_{k,t}\to 0$ is anchored to its previous state and suppressed in the decoder, so it neither absorbs the occluder nor explains unrelated content, while its representation stays available for reactivation. This inactive-but-persistent regime is the one prior work lacks.

\begin{table}[t]
\centering
\caption{\textbf{Conceptual comparison of slot-control mechanisms.}
Methods: SAVi~\citep{kipf2021conditional}, VideoSAUR~\citep{zadaianchuk2023object}, SlotContrast~\citep{manasyan2025temporally}, RandSF.Q~\citep{zhao2026predicting}, SmoothSA~\citep{zhao2026smoothsa}, SOLV~\citep{aydemir2023self}, AdaSlot~\citep{fan2024adaptive}, and MetaSlot~\citep{liu2026metaslot}.
Pathway I/II: whether an absent object's slot is protected from state updates / removed from decoding (Sec.~\ref{sec:problem}).
\cmark~yes; \xmark~no; \na~not applicable.
$^\dagger$Image-level module; evaluated on video only by plugging it into VideoSAUR.}
\label{tab:conceptual}
\resizebox{\linewidth}{!}{%
\footnotesize
\setlength{\tabcolsep}{2pt}
\begin{tabular}{l c c l c c c c c l}
\toprule
& & \multicolumn{2}{c}{\textbf{Participation decision}}
  & \multicolumn{3}{c}{\textbf{Gated pathways}}
  & \multicolumn{2}{c}{\textbf{Slot lifecycle}} & \\
\cmidrule(lr){3-4}\cmidrule(lr){5-7}\cmidrule(lr){8-9}
\textbf{Method} & \textbf{Video}
  & \makecell{Per-slot,\\per-frame} & \makecell[l]{Conditioned on}
  & \makecell{Pathway I\\(state)} & \makecell{Pathway II\\(decoder)} & \makecell{Shared\\control}
  & \makecell{Persistent\\when inactive} & \makecell{Reversible}
  & \makecell[l]{Slot under full \\ occlusion} \\
\midrule
\multicolumn{10}{l}{\textit{Slot propagation}} \\
SAVi         & \cmark & \xmark & \na & \xmark & \xmark & \xmark & \xmark & \na & Updated \& decoded \\
VideoSAUR    & \cmark & \xmark & \na & \xmark & \xmark & \xmark & \xmark & \na & Updated \& decoded \\
SlotContrast & \cmark & \xmark & \na & \xmark & \xmark & \xmark & \xmark & \na & Updated \& decoded \\
RandSF.Q     & \cmark & \xmark & \na & \xmark & \xmark & \xmark & \xmark & \na & Updated \& decoded \\
SmoothSA     & \cmark & \xmark & \na & \xmark & \xmark & \xmark & \xmark & \na & Updated \& decoded \\
\midrule
\multicolumn{10}{l}{\textit{Slot pruning / merging}} \\
SOLV     & \cmark & \cmark & Inter-slot similarity & \xmark & \xmark & \xmark & \xmark & \xmark & Merged \\
AdaSlot  & \xmark & \cmark & Frame content         & \na    & \cmark & \na    & \xmark & \na    & Dropped \\
MetaSlot & \xmark$^\dagger$ & \cmark & Frame content & \na & \cmark & \na & \xmark & \na & Dropped \\
\midrule
\multicolumn{10}{l}{\textit{Slot lifecycle}} \\
\rowcolor{tsarow}
\textbf{TSA (ours)} & \cmark & \cmark & \textbf{Candidate + memory}
 & \cmark & \cmark & \cmark & \cmark & \cmark & \textbf{Frozen \& silenced} \\
\bottomrule
\end{tabular}}
\end{table}

\section{Dataset details}
\label{app:datasets}
\begin{figure}[t]
\centering
\newcommand{\rowlabel}[2]{%
  \raisebox{#1\height}{\rotatebox{90}{\small \textbf{#2}}}%
}
\begin{tabular}{c@{\hspace{2pt}}c@{\hspace{2pt}}c@{\hspace{2pt}}c@{\hspace{2pt}}c}
\rowlabel{0.7}{MOVi-C} &
\includegraphics[width=0.22\linewidth]{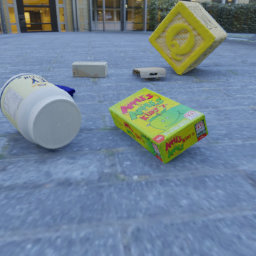} &
\includegraphics[width=0.22\linewidth]{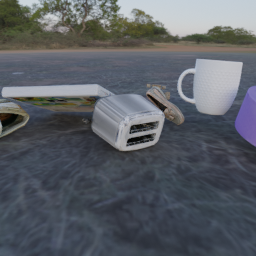} &
\includegraphics[width=0.22\linewidth]{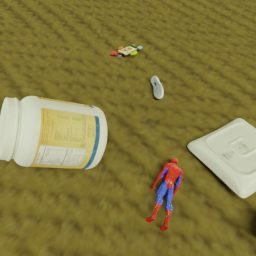} &
\includegraphics[width=0.22\linewidth]{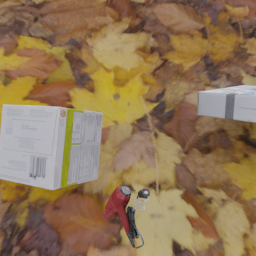} \\[2pt]

\rowlabel{0.8}{MOVi-E} &
\includegraphics[width=0.22\linewidth]{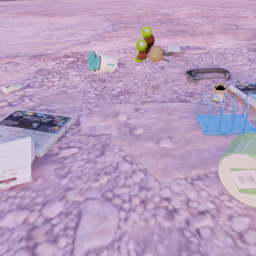} &
\includegraphics[width=0.22\linewidth]{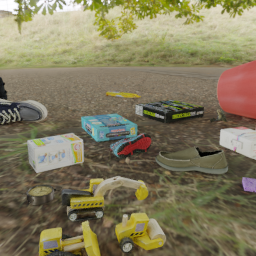} &
\includegraphics[width=0.22\linewidth]{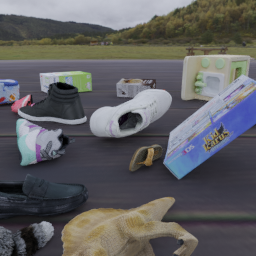} &
\includegraphics[width=0.22\linewidth]{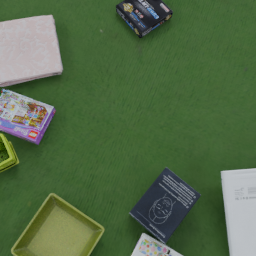} \\[2pt]

\rowlabel{0.15}{YouTube-VIS HQ} &
\includegraphics[width=0.22\linewidth]{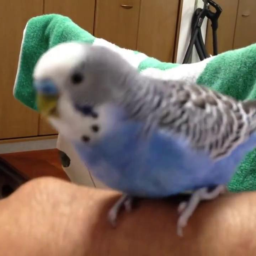} &
\includegraphics[width=0.22\linewidth]{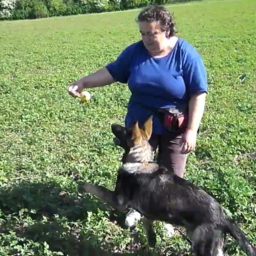} &
\includegraphics[width=0.22\linewidth]{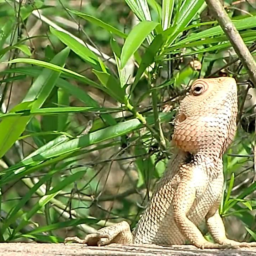} &
\includegraphics[width=0.22\linewidth]{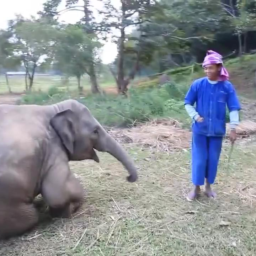} \\[2pt]

\rowlabel{1.4}{OVIS} &
\includegraphics[width=0.22\linewidth]{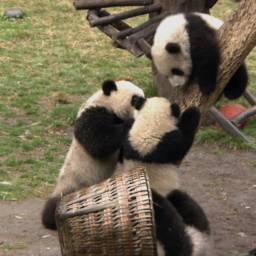} &
\includegraphics[width=0.22\linewidth]{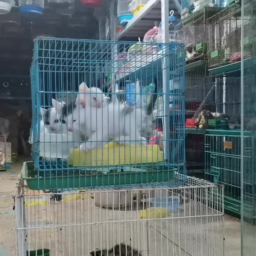} &
\includegraphics[width=0.22\linewidth]{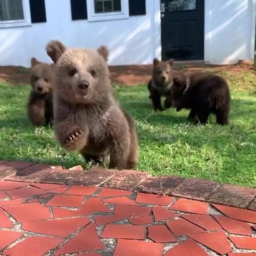} &
\includegraphics[width=0.22\linewidth]{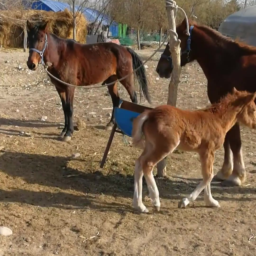} \\
\end{tabular}
\caption{\textbf{Qualitative samples from the four benchmarks.} Each row shows four representative videos from a single dataset, illustrating the visual diversity within each benchmark. MOVi-C and MOVi-E provide controlled synthetic scenes with known dynamics; YouTube-VIS HQ contributes natural appearance variation and object motion; OVIS contains crowded scenes with severe occlusion and long object trajectories.
}
\label{fig:dataset_samples}
\end{figure}

We evaluate our approach on four complementary video benchmarks that span synthetic and real-world domains, ranging from controlled multi-object scenes to crowded videos with severe occlusion. Table~\ref{tab:dataset_comparison} summarizes the key characteristics of each benchmark, while Fig.~\ref{fig:dataset_samples} shows qualitative examples illustrating the visual diversity and difficulty of each dataset.

\paragraph{MOVi-C and MOVi-E.}
MOVi-C and MOVi-E~\citep{greff2022kubric} are synthetic multi-object video benchmarks generated with the Kubric simulator. They serve as controlled settings for evaluating object-centric video grouping under known object dynamics. Both datasets consist of rigid objects with stable appearance, but differ substantially in scene complexity: MOVi-C features moderately cluttered scenes on textured backgrounds, whereas MOVi-E exhibits denser object layouts, stronger camera motion, and more frequent inter-object occlusions.

\paragraph{YouTube-VIS HQ.}
YouTube-VIS HQ~\citep{ke2022video} is a real-world video instance segmentation benchmark with high-quality, manually refined object mask annotations. It contains natural videos featuring non-rigid objects with substantial appearance variation, diverse motion patterns, and cluttered backgrounds, making it well suited for evaluating grouping performance in unconstrained settings.

\paragraph{OVIS.}
OVIS~\citep{qi2022occluded} is a real-world video instance segmentation benchmark explicitly designed around heavy occlusion. It contains crowded scenes with non-rigid objects, long object trajectories, and frequent partial or full visibility changes, providing a rigorous stress test for temporal persistence and object re-identification under severe occlusion.

\begin{table}[t]
\caption{Comparison of the four video benchmarks used in our evaluation.}
\label{tab:dataset_comparison}
\begin{center}
\setlength{\tabcolsep}{8pt}
\begin{tabular}{lccc}
\toprule
\textbf{Dataset} & \textbf{Domain} & \textbf{Object type} & \textbf{Main challenge} \\
\midrule
MOVi-C         & Synthetic & Rigid     & Moderate clutter                    \\
MOVi-E         & Synthetic & Rigid     & Dense scenes, camera motion         \\
YouTube-VIS HQ & Real      & Non-rigid & Natural appearance variation        \\
OVIS           & Real      & Non-rigid & Severe occlusion, long trajectories \\
\bottomrule
\end{tabular}
\end{center}
\end{table}

\section{Evaluation metrics}
\label{app:eval}

\paragraph{Foreground adjusted Rand index (FG-ARI).}
ARI measures the agreement between two clustering assignments over a set of elements, corrected for chance agreement. Following the standard object-centric protocol~\citep{locatello2020object,kipf2021conditional}, we restrict the computation to ground-truth foreground pixels by excluding the background label, which isolates the signal of the metric to how well distinct objects are separated from one another rather than from the scene background. We compute FG-ARI at the video level by treating all pixels across frames of a sequence as a single clustering problem, so that the metric reflects not only segmentation quality within frames but also identity consistency of slot assignments over time.

\paragraph{Mean best overlap (mBO).}
mBO quantifies per-object mask coverage by selecting, for each ground-truth instance, the predicted slot mask with the highest intersection-over-union (IoU), and averaging these best-match IoUs across all instances and sequences. Unlike FG-ARI, mBO retains background pixels in the IoU computation, which makes it more sensitive to mask boundary precision and to spurious slot activations on non-object regions.

\paragraph{Higher order tracking accuracy (HOTA).}
HOTA~\citep{luiten2021hota} jointly measures detection accuracy and association quality, avoiding the bias toward either detection or tracking that arises in single-metric evaluations.
We compute HOTA with the official TrackEval implementation ~\citep{luiten2020trackeval}, treating each slot index as a tracker identity, without access to identity labels at training time.

\paragraph{IDF1.}
IDF1~\citep{ristani2016performance} measures the ratio of correctly identified detections over the mean of ground-truth and computed detections, using identity-consistent matching.
Each predicted slot is matched to a ground-truth instance by a single sequence-level assignment, following the official TrackEval implementation.

\paragraph{Seeding and averaging.}
Results on the main benchmarks and the downstream evaluations are reported as mean $\pm$ std
over 3 independent random seeds controlling model initialization and data ordering. Ablation
and sensitivity studies are run with a single seed, shared across all conditions, so that
the conditions are compared under the same random seed and data ordering.

\section{Implementation details}
\label{app:impl}
\paragraph{Model.}
We use a frozen DINOv2 ViT-S/14~\citep{oquab2023dinov2} as the visual encoder. Each frame is resized from $256\times256$ to $224\times224$, producing $N=256$ patch tokens with feature dimension $d_f=384$. The patch features are projected by a two-layer MLP before being used as keys and values in Slot Attention. Slot Attention uses $K$ slots (dataset-specific, see Table~\ref{tab:hyperparams}) with slot dimension $d=256$, and runs for $3$ iterations on the first frame of each clip and $1$ iteration on subsequent frames ~\citep{zhao2026smoothsa}. The temporal query transitioner $T_\phi$ follows the RandSF.Q transition module~\citep{zhao2026predicting}, instantiated as a single Transformer decoder layer with $4$ attention heads. The Slot Activation Estimator $\Phi_{\mathrm{act}}$ is a two-layer MLP with ReLU activations that maps the concatenation of the current slot candidate ($d=256$) and the previous temporal memory $\vm_{k,t-1}$ ($d_h=64$) to a scalar activation logit. The Temporal Context Encoder $\Psi_{\mathrm{tce}}$ is a single-layer GRU shared across slots, with hidden dimension $d_h=64$, that produces $\vm_{k,t}$ from the post-gate slot state at each step and is reset at the start of each video. The temporal memory $\vm_{k,t}$ is consumed only by $\Phi_{\mathrm{act}}$ and does not feed back into the slot representation directly. The decoder is an autoregressive Transformer with model dimension $d_f=384$ that reconstructs DINOv2 features rather than RGB pixels~\citep{zadaianchuk2023object,manasyan2025temporally}. The activation log-bias from Eq.~\ref{eq:log_bias} is added to the decoder cross-attention logits at each layer.

\paragraph{Training.}
We train the model with Adam~\citep{kingma2014adam} for $50{,}000$ steps using a batch size of $8$ video clips. Each training sample is a contiguous segment of length $T$ frames (dataset-specific; see Table~\ref{tab:hyperparams}). The learning rate is initialized to $5\times10^{-5}$, linearly warmed up over the first $2{,}500$ steps, and the gradient norm is clipped at $0.05$. The total loss is $\Ls = \Ls_{\mathrm{recon}} + \lambda_{\mathrm{ssc}}\Ls_{\mathrm{ssc}} + \lambda_{\mathrm{reg}}(t)\,\Ls_{\mathrm{reg}}$, where $\lambda_{\mathrm{ssc}}=0.5$ throughout and the regularization weight $\lambda_{\mathrm{reg}}(t)$ follows a two-stage schedule: it is held at zero for the first $T_{\mathrm{warmup}}$ steps, then linearly increased to its target value over the next $T_{\mathrm{ramp}}$ steps.

\paragraph{Inference.}
At test time, slot activations $\evalpha_{k,t} \in (0,1)$ remain continuous and are not thresholded; the same activation-gated state update and activation log-bias used during training are applied unchanged. Full videos are processed sequentially, with slot states and per-slot temporal memories $\vm_{k,t}$ propagated across frames without resetting.

\paragraph{Hardware.}
All experiments are run on a single NVIDIA RTX A6000 GPU.

Table~\ref{tab:hyperparams} summarizes the dataset-specific implementation details, including the number of slots $K$, training segment length $T$, loss coefficients, and regularization schedules.

\begin{table}[t]
\caption{Implementation details across the four benchmarks. Shared hyperparameters are listed once across all columns; dataset-specific values are given per benchmark.}
\label{tab:hyperparams}
\begin{center}
\setlength{\tabcolsep}{6pt}
\begin{tabular}{lcccc}
\toprule
\multirow{2}{*}{\textbf{Hyperparameter}}
  & \multicolumn{4}{c}{\textbf{Benchmarks}} \\
\cmidrule(lr){2-5}
  & \textbf{MOVi-C} & \textbf{MOVi-E} & \textbf{YouTube-VIS HQ} & \textbf{OVIS} \\
\midrule
\multicolumn{5}{l}{\emph{Optimization}} \\
\quad Optimizer                             & \multicolumn{4}{c}{Adam} \\
\quad Training steps                        & \multicolumn{4}{c}{$50{,}000$} \\
\quad Batch size (clips)                    & \multicolumn{4}{c}{$8$} \\
\quad Training segment length $T$           & $6$ & $6$ & $5$ & $10$ \\
\quad Initial learning rate                 & \multicolumn{4}{c}{$5\times10^{-5}$} \\
\quad LR warm-up steps                      & \multicolumn{4}{c}{$2{,}500$} \\
\quad Gradient norm clip                    & \multicolumn{4}{c}{$0.05$} \\
\midrule
\multicolumn{5}{l}{\emph{Visual encoder (frozen)}} \\
\quad Backbone                              & \multicolumn{4}{c}{DINOv2 ViT-S/14} \\
\quad Input resolution                      & \multicolumn{4}{c}{$256 \to 224$} \\
\quad Number of image tokens $N$            & \multicolumn{4}{c}{$256$} \\
\quad Feature dimension $d_f$               & \multicolumn{4}{c}{$384$} \\
\midrule
\multicolumn{5}{l}{\emph{Slot Attention}} \\
\quad Number of slots $K$                   & $11$ & $24$ & $7$ & $22$ \\
\quad Slot dimension $d$                    & \multicolumn{4}{c}{$256$} \\
\quad Key / value dimension                 & \multicolumn{4}{c}{$384$} \\
\quad FFN dimension                         & \multicolumn{4}{c}{$1{,}024$} \\
\quad Iterations (first / subsequent frame) & \multicolumn{4}{c}{$3\;/\;1$} \\
\midrule
\multicolumn{5}{l}{\emph{Temporal Query Transitioner $T_\phi$}} \\
\quad Type                                  & \multicolumn{4}{c}{Transformer decoder layer} \\
\quad Heads / FFN dimension                 & \multicolumn{4}{c}{$4\;/\;1{,}024$} \\
\quad Dropout                               & \multicolumn{4}{c}{$0.5$} \\
\midrule
\multicolumn{5}{l}{\emph{Slot Activation Estimator $\Phi_{\mathrm{act}}$}} \\
\quad Type                                  & \multicolumn{4}{c}{Two-layer MLP (ReLU)} \\
\quad Input dim ($d + d_h$) / hidden dim    & \multicolumn{4}{c}{$320\;/\;128$} \\
\midrule
\multicolumn{5}{l}{\emph{Temporal Context Encoder $\Psi_{\mathrm{tce}}$}} \\
\quad Type                                  & \multicolumn{4}{c}{Single-layer GRU} \\
\quad Input dim $d$ / hidden dim $d_h$      & \multicolumn{4}{c}{$256\;/\;64$} \\
\midrule
\multicolumn{5}{l}{\emph{Decoder}} \\
\quad Type                                  & \multicolumn{4}{c}{Autoregressive Transformer decoder} \\
\quad Layers / heads / FFN dim              & \multicolumn{4}{c}{$4\;/\;4\;/\;1{,}536$} \\
\quad Model dimension $d_f$                 & \multicolumn{4}{c}{$384$} \\
\quad Reconstruction target                 & \multicolumn{4}{c}{DINOv2 patch features} \\
\midrule
\multicolumn{5}{l}{\emph{Loss coefficients and schedule}} \\
\quad $\lambda_{\mathrm{ssc}}$              & \multicolumn{4}{c}{$0.5$} \\
\quad $\lambda_{\mathrm{reg}}$              & $0.06$    & $0.02$    & $0.10$    & $0.15$    \\
\quad $\beta$                               & $1$    & $3$    & $1.5$   & $1$   \\
\quad $T_{\mathrm{warmup}}$ (steps)         & \multicolumn{4}{c}{$1{,}000$} \\
\quad $T_{\mathrm{ramp}}$ (steps)           & 
\multicolumn{4}{c}{$6{,}000$} \\
\bottomrule
\end{tabular}
\end{center}
\end{table}

\section{Sensitivity to regularization hyperparameters}
\label{app:sensitivity}
All components inherited from prior work are kept at their published settings
(Table~\ref{tab:hyperparams}). TSA introduces two new loss weights, both in
$\Ls_{\mathrm{reg}} = \Ls_{\mathrm{usage}} + \beta\,\Ls_{\mathrm{sparse}}$ of Eq.~\ref{eq:activation_reg}:
$\lambda_{\mathrm{reg}}$, which sets how strongly the regularizer competes with reconstruction, and
$\beta$, which weights $\Ls_{\mathrm{sparse}}$ against $\Ls_{\mathrm{usage}}$; the warm-up schedule
of $\lambda_{\mathrm{reg}}$ is shared across benchmarks. We sweep one weight at a time on
YouTube-VIS HQ, keeping the other at its paper value for this benchmark
($\lambda_{\mathrm{reg}} = 0.10$, $\beta = 1.5$; Table~\ref{tab:hyperparams}).

\paragraph{Effect of $\lambda_{\mathrm{reg}}$.}
Table~\ref{tab:sensitivity}~(left) shows that, on YouTube-VIS HQ, performance is best at
intermediate values and degrades at both ends of the range. A weak regularizer exerts less
pressure to deactivate slots, while a strong one increasingly outweighs the reconstruction
objective. On this benchmark, performance is stable between $0.10$ and $0.15$, where every
metric stays within $2.3$ points.

\paragraph{Effect of $\beta$.}
Table~\ref{tab:sensitivity}~(right) shows that, on YouTube-VIS HQ, performance drops mainly when
$\beta$ is too small, where intermediate activations are penalized less. Performance improves as
$\beta$ increases, and for $\beta \geq 1.5$ every metric stays within $1.8$ points of the paper
setting.

These ranges are specific to YouTube-VIS HQ: the appropriate values depend on the benchmark, and
the settings used for the other benchmarks differ (Table~\ref{tab:hyperparams}).

\begin{table}[t]
\caption{Sensitivity to the two regularization hyperparameters on YouTube-VIS HQ.
\emph{Left:} regularization strength, varying $\lambda_{\mathrm{reg}}$ with $\beta = 1.5$.
\emph{Right:} sparsity weight, varying $\beta$ with $\lambda_{\mathrm{reg}} = 0.10$.
Best per column in \textbf{bold}; the setting used in the paper is marked $\dagger$. Results from a single seed.}
\label{tab:sensitivity}
\begin{center}
\setlength{\tabcolsep}{5pt}
\begin{minipage}[t]{0.48\linewidth}
\centering
\begin{tabular}{ccccc}
\toprule
$\lambda_{\mathrm{reg}}$ & FG-ARI$\uparrow$ & mBO$\uparrow$ & HOTA$\uparrow$ & IDF1$\uparrow$ \\
\midrule
$0.05$ & 72.3 & 49.3 & 37.3 & 38.4 \\
$0.10^{\dagger}$ & $\mathbf{76.3}$ & 55.8 & 46.3 & $\mathbf{49.2}$ \\
$0.15$ & 74.0 & $\mathbf{56.8}$ & $\mathbf{46.5}$ & 48.9 \\
$0.20$ & 73.6 & 52.7 & 40.9 & 40.8 \\
$0.25$ & 68.4 & 47.4 & 33.1 & 30.5 \\
\bottomrule
\end{tabular}
\end{minipage}
\hfill
\begin{minipage}[t]{0.48\linewidth}
\centering
\begin{tabular}{ccccc}
\toprule
$\beta$ & FG-ARI$\uparrow$ & mBO$\uparrow$ & HOTA$\uparrow$ & IDF1$\uparrow$ \\
\midrule
$0.5$ & 74.2 & 51.5 & 39.5 & 40.7 \\
$1.0$ & 75.3 & 53.7 & 42.3 & 44.6 \\
$1.5^{\dagger}$ & $\mathbf{76.3}$ & 55.8 & 46.3 & 49.2 \\
$2.0$ & 75.7 & $\mathbf{57.6}$ & $\mathbf{48.0}$ & $\mathbf{50.9}$ \\
$2.5$ & 75.8 & 55.3 & 45.7 & 48.6 \\
\bottomrule
\end{tabular}
\end{minipage}
\end{center}
\end{table}

\section{Additional ablation study on YouTube-VIS HQ}
\label{app:ablation_ytvis}
\noindent \textbf{A. Activation-gated pathways.}
Table~\ref{tab:ablation_ytvis}~(left): gating the decoder alone (Exp.~\#2) leaves the metrics nearly unchanged
over the baseline, gating the state update alone (Exp.~\#3) gives a large improvement, and using both (Exp.~\#4) is best on all three metrics. Decoder gating contributes most when combined
with state gating, adding $+9.6$ HOTA over Exp.~\#3.

\noindent \textbf{B. Regularization.}
Table~\ref{tab:ablation_ytvis}~(middle): $\Ls_{\mathrm{sparse}}$ alone (Exp.~\#2) stays close to
reconstruction alone (Exp.~\#1), $\Ls_{\mathrm{usage}}$ (Exp.~\#3) gives a large gain, and using
both (Exp.~\#4) is best on all three metrics, adding a further $+10.0$ HOTA over Exp.~\#3.

\noindent \textbf{C. Temporal memory.}
Table~\ref{tab:ablation_ytvis}~(right): the accumulated memory $\vm_{k,t-1}$ (Exp.~\#3) is best on
mBO and HOTA, followed by the previous slot state $\vs_{k,t-1}$ (Exp.~\#2) and no temporal context
(Exp.~\#1); the three are within $1.3$ FG-ARI of each other, with Exp.~\#2 marginally ahead. The
gap between the three is mostly in HOTA ($4.2$, against $1.3$ FG-ARI and $3.5$ mBO), since
YouTube-VIS HQ has shorter sequences and fewer long occlusions than OVIS.

\begin{table}[t]
\caption{Additional ablation studies on YouTube-VIS HQ. \emph{Left:} activation-gated pathways. \emph{Middle:} regularization terms in $\Ls_{\mathrm{reg}}$. \emph{Right:} temporal memory. Results from a single seed. Best per column in \textbf{bold}.}
\label{tab:ablation_ytvis}
\begin{center}
\setlength{\tabcolsep}{4pt}
\renewcommand{\arraystretch}{1.15}
\begin{minipage}[t]{0.36\linewidth}
\centering
\resizebox{\linewidth}{!}{%
\begin{tabular}{cccccc}
\toprule
\multirow{2}{*}{\textbf{Exp.}} & \textbf{State} & \textbf{Decoder}
& \multicolumn{3}{c}{\textbf{Metrics}} \\
\cmidrule(lr){2-3}\cmidrule(lr){4-6}
 & \textbf{update} & \textbf{participate}
 & FG-ARI$\uparrow$ & mBO$\uparrow$ & HOTA$\uparrow$ \\
\midrule
\#1 & \ding{55} & \ding{55} & 57.1 & 38.0 & 21.8 \\
\#2 & \ding{55} & \ding{51} & 58.8 & 37.7 & 21.9 \\
\#3 & \ding{51} & \ding{55} & 74.8 & 51.9 & 36.7 \\
\textbf{\#4} & \ding{51} & \ding{51}
 & \textbf{76.3} & \textbf{55.8} & \textbf{46.3} \\
\bottomrule
\end{tabular}}
\end{minipage}
\hfill
\begin{minipage}[t]{0.32\linewidth}
\centering
\resizebox{\linewidth}{!}{%
\begin{tabular}{cccccc}
\toprule
\multirow{2}{*}{\textbf{Exp.}} & \multicolumn{2}{c}{\textbf{Loss}}
& \multicolumn{3}{c}{\textbf{Metrics}} \\
\cmidrule(lr){2-3}\cmidrule(lr){4-6}
 & $\Ls_{\mathrm{sparse}}$ & $\Ls_{\mathrm{usage}}$
 & FG-ARI$\uparrow$ & mBO$\uparrow$ & HOTA$\uparrow$ \\
\midrule
\#1 & \ding{55} & \ding{55} & 58.6 & 41.3 & 24.0 \\
\#2 & \ding{51} & \ding{55} & 57.6 & 38.8 & 22.6 \\
\#3 & \ding{55} & \ding{51} & 72.2 & 50.5 & 36.3 \\
\textbf{\#4} & \ding{51} & \ding{51}
 & \textbf{76.3} & \textbf{55.8} & \textbf{46.3} \\
\bottomrule
\end{tabular}}
\end{minipage}
\hfill
\begin{minipage}[t]{0.28\linewidth}
\centering
\resizebox{\linewidth}{!}{%
\begin{tabular}{ccccc}
\toprule
\multirow{2}{*}{\textbf{Exp.}} & \multirow{2}{*}{\shortstack{\textbf{Temporal}\\\textbf{memory}}}
& \multicolumn{3}{c}{\textbf{Metrics}} \\
\cmidrule(lr){3-5}
 & & FG-ARI$\uparrow$ & mBO$\uparrow$ & HOTA$\uparrow$ \\
\midrule
\#1 & \ding{55} & 76.2 & 52.3 & 42.1 \\
\#2 & $\vs_{k,t-1}$ & \textbf{77.5} & 54.4 & 45.1 \\
\textbf{\#3} & $\vm_{k,t-1}$
 & {76.3} & \textbf{55.8} & \textbf{46.3} \\
\bottomrule
\end{tabular}}
\end{minipage}
\end{center}
\end{table}

\section{Additional analysis and downstream task evaluation}
\label{app:downstream}

\subsection{Activation distribution under each regularization term}
\label{app:activation_hist}
\begin{figure}[t]
    \centering
    \includegraphics[width=\linewidth]{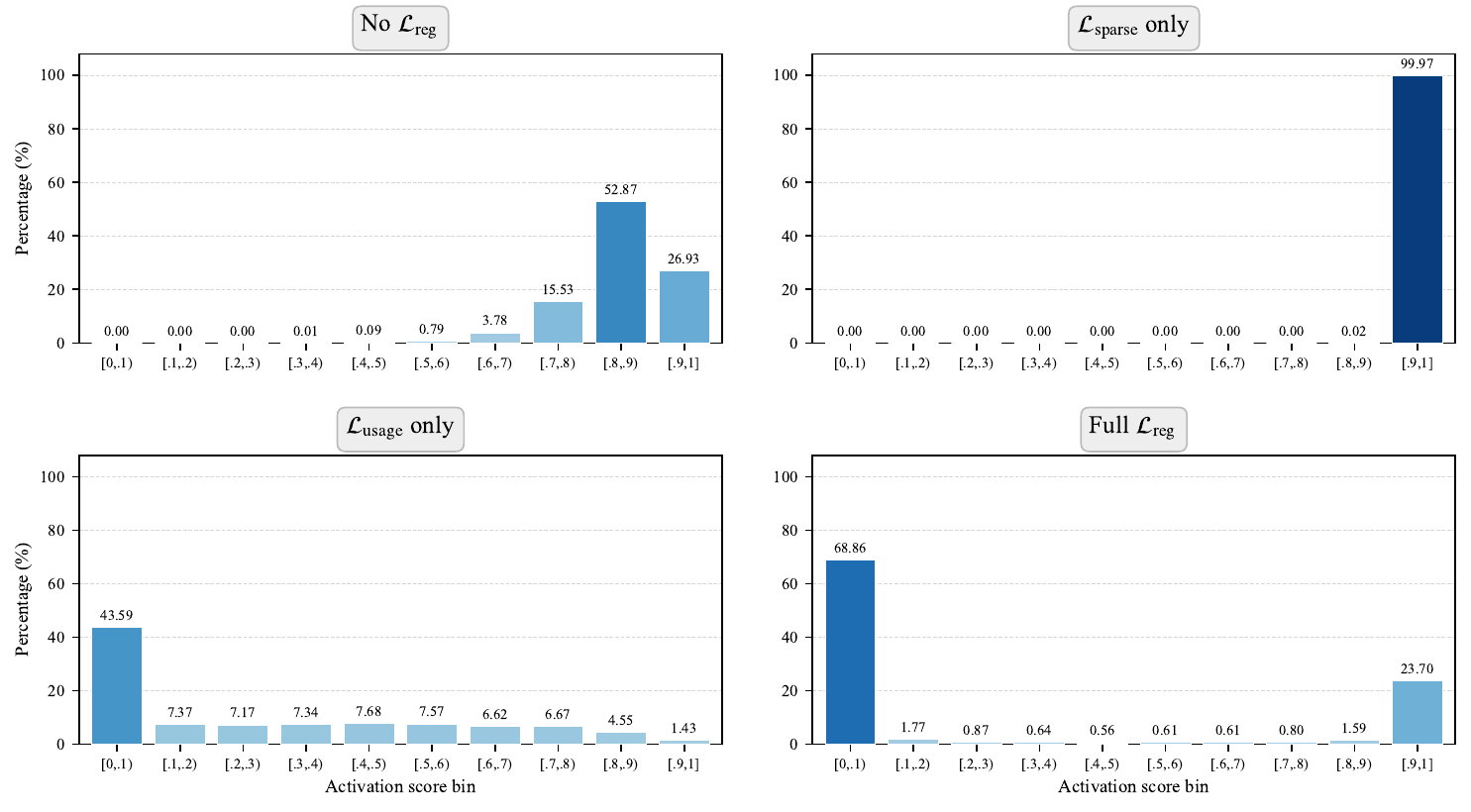}
    \caption{\textbf{Distribution of activation scores $\evalpha_{k,t}$ on YouTube-VIS HQ} under
    the four regularization settings of Table~\ref{tab:ablation_ytvis}~(middle). Each panel is a
    10-bin histogram over $[0,1]$, reported as \% of all slot--frame pairs.}
    \label{fig:activation_hist}
\end{figure}

Table~\ref{tab:ablation_ytvis}~(middle) reports what each term of $\Ls_{\mathrm{reg}}$
contributes in metrics. Here we look at what it does to the activation score itself. For the
same four settings, we collect all per-slot, per-frame activations $\evalpha_{k,t}$ on
YouTube-VIS HQ and bin them into ten intervals (Fig.~\ref{fig:activation_hist}).

Without $\Ls_{\mathrm{reg}}$, nearly all activations lie above $0.6$ and about $80\%$ above
$0.8$: every slot stays active at every frame, which is the full-activation collapse described
in Sec.~\ref{subsec:objectives} and leaves TSA behaving like unconditional propagation. Adding
$\Ls_{\mathrm{sparse}}$ alone does not break this collapse but sharpens it, moving nearly
all activations into the last bin. This is expected, since $\Ls_{\mathrm{sparse}}$ only
pushes activations away from intermediate values and never gives a slot a reason to
deactivate, which is why it stays at the level of the unregularized model.

$\Ls_{\mathrm{usage}}$ alone does move a large fraction of activations below $0.1$, so slots
now genuinely deactivate. The remaining activations, however, are spread almost evenly over
the intermediate bins. A slot in this range is neither properly frozen nor properly
silenced: it is partially overwritten by the current frame while still contributing to
decoding, so it behaves as a mixture of an active and an inactive slot rather than committing
to either. This is the ambiguous-gating regime discussed in Sec.~\ref{subsec:objectives}.

With both terms, the distribution becomes clearly bimodal: the intermediate bins are largely
emptied and the activations concentrate at the two ends. Each slot is thus either kept
inactive but persistent, or allowed to track and reconstruct normally, with few frames left
in between. The two terms are complementary in exactly the way the ablation suggests:
$\Ls_{\mathrm{usage}}$ makes deactivation happen at all, while $\Ls_{\mathrm{sparse}}$ makes
each decision decisive. Because the learned activations are already close to binary, they
can be used continuously at inference without any thresholding.

\subsection{Representation drift across occlusion intervals}
\label{app:drift}

To directly verify that the activation-gated state update mitigates the update-induced state drift discussed in Sec.~\ref{sec:problem}, we measure the \emph{representation drift} of a slot across each occlusion interval as the per-dimension mean squared difference between its states:
\begin{equation}
d_{\mathrm{drift}}(k) = \frac{1}{d} \left\| \vs_{k,t_{\mathrm{post}}} - \vs_{k,t_{\mathrm{pre}}} \right\|^2,
\label{eq:occlusion_drift}
\end{equation}
where $t_{\mathrm{pre}}$ denotes the last visible frame before an object becomes fully occluded, and $t_{\mathrm{post}}$ denotes the first visible frame after the object reappears. For each object that becomes fully occluded and later reappears, we identify the predicted slot tracking it by selecting the slot with the highest mask IoU at $t_{\mathrm{pre}}$. We then compute $d_{\mathrm{drift}}$ between the pre-occlusion and post-reappearance states of that slot. An object is absent in any frame where it has no ground-truth mask, and one occlusion interval is a run of absent frames between two frames where the object is annotated.

\begin{figure}[t]
    \centering
    \includegraphics[width=\linewidth]{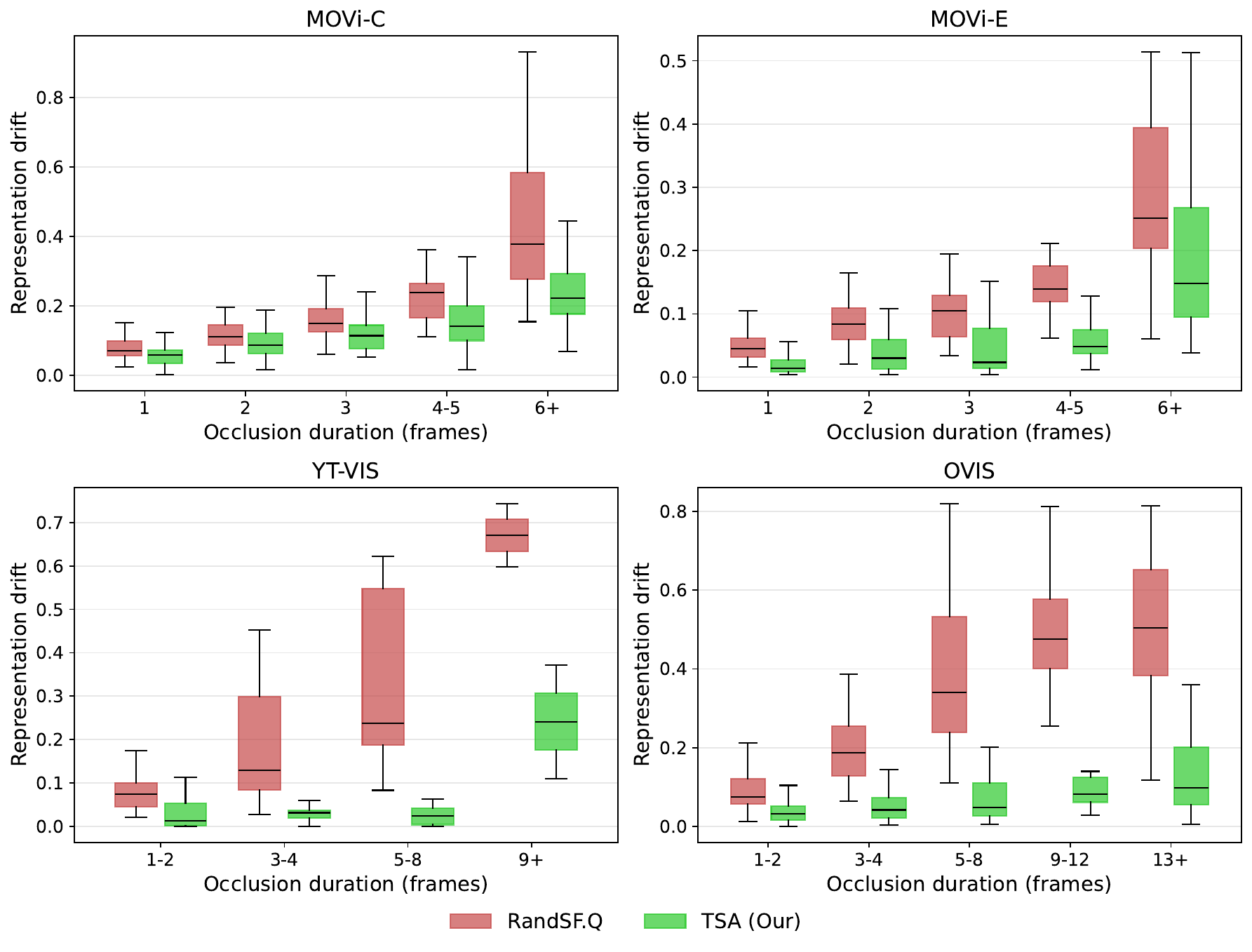}
    \caption{\textbf{Representation drift across occlusion intervals.} Box plots show the distribution of the per-dimension representation drift $d_{\mathrm{drift}}$ across occlusion-duration bins on MOVi-C, MOVi-E, YouTube-VIS HQ, and OVIS.}
    \label{fig:drift}
\end{figure}

We compare RandSF.Q, an unconditional propagation baseline that updates every slot at every step, against TSA. Fig.~\ref{fig:drift} shows the distribution of $d_{\mathrm{drift}}$ stratified by occlusion duration on MOVi-C, MOVi-E, YouTube-VIS HQ, and OVIS.

Two observations stand out. First, representation drift generally increases with occlusion duration, indicating that longer absences make it more difficult to preserve the pre-occlusion slot identity. Second, TSA consistently produces lower drift than RandSF.Q across all datasets and duration bins,
with the gap widening at long durations. These results provide direct evidence that the activation-gated state update of Eq.~\ref{eq:state_gate} preserves slot identity more effectively across occlusions by anchoring inactive slots to their previous states, consistent with the cumulative-drift analysis in Eq.~\ref{eq:cumulative_drift}.

\subsection{Downstream task evaluation}
\label{app:downstream_tasks}

To further assess the quality of the slot representations learned by TSA, we evaluate them on two downstream tasks on YouTube-VIS HQ. Both tasks operate on \emph{frozen} slot representations, isolating the contribution of the representation itself from any task-specific finetuning. We compare TSA against SlotContrast~\citep{manasyan2025temporally}, RandSF.Q~\citep{zhao2026predicting}, and SmoothSA~\citep{zhao2026smoothsa} under identical training and evaluation protocols. The two tasks probe complementary properties of the representation: object recognition is a per-frame appearance test, whereas dynamics prediction is a cross-frame temporal-stability test.

\paragraph{Object recognition.}
Following RandSF.Q~\citep{zhao2026predicting}, we freeze the object-centric model and train a two-layer MLP to predict the object class and bounding box corresponding to each slot, supervised by the object class labels and bounding box annotations in the dataset. Each predicted slot is matched to a ground-truth instance using a first-visible-frame majority-overlap rule. This task probes whether slot representations preserve discriminative per-frame object information, including semantic category and spatial localization. We report Top-1 and Top-3 classification accuracy, bounding-box IoU, and the number of matched samples.

\paragraph{Object dynamics prediction.}
Following SlotContrast~\citep{manasyan2025temporally}, we train SlotFormer~\citep{wu2022slotformer} on top of the frozen slot representations to predict object dynamics. SlotFormer receives $10$ burn-in frames of inferred slots and autoregressively predicts slots for $5$ rollout steps. Both the object-centric model and SlotFormer operate entirely in feature space, and SlotFormer is trained using only the slot reconstruction loss. Unlike object recognition, which is evaluated on matched per-frame slots, dynamics prediction depends strongly on whether slot identities and trajectories remain stable over time. We therefore use this task to assess the temporal consistency and predictability of the learned slot representations. We report FG-ARI and mBO on the predicted slot rollouts.

\begin{table}[t]
\caption{Object recognition on YouTube-VIS HQ. A two-layer MLP is trained on frozen slot representations. Mean $\pm$ std over 3 seeds.}
\label{tab:ytvis_recognition}
\begin{center}
\setlength{\tabcolsep}{6pt}
\begin{tabular}{lcccc}
\toprule
\textbf{Method} & \textbf{Top-1}$\uparrow$ & \textbf{Top-3}$\uparrow$ & \textbf{Bbox IoU}$\uparrow$ & \textbf{Matched} \\
\midrule
SlotContrast + MLP  & $85.8_{\pm0.3}$ & $95.8_{\pm0.4}$ & $\mathbf{51.5_{\pm0.5}}$ & ${9249_{\pm41}}$ \\
RandSF.Q + MLP      & $90.5_{\pm0.3}$ & $\mathbf{97.9_{\pm0.3}}$ & $50.6_{\pm0.4}$ & $8979_{\pm123}$ \\
SmoothSA + MLP      & $90.4_{\pm0.2}$ & $97.6_{\pm0.1}$ & $42.6_{\pm1.4}$ & $8957_{\pm34}$ \\
\textbf{TSA (ours) + MLP} & $\mathbf{91.3_{\pm0.9}}$ & $\mathbf{97.9_{\pm0.3}}$ & $\mathbf{51.5_{\pm0.2}}$ & $7908_{\pm36}$ \\
\bottomrule
\end{tabular}
\end{center}
\end{table}

\begin{table}[t]
\caption{Object dynamics prediction on YouTube-VIS HQ. Mean $\pm$ std over 3 seeds.}
\label{tab:ytvis_dynamics}
\begin{center}
\setlength{\tabcolsep}{8pt}
\begin{tabular}{lcc}
\toprule
\textbf{Method} & \textbf{FG-ARI}$\uparrow$ & \textbf{mBO}$\uparrow$ \\
\midrule
SlotContrast        & $29.5_{\pm0.2}$ & $33.2_{\pm0.1}$ \\
RandSF.Q            & $38.2_{\pm0.5}$ & $43.7_{\pm0.6}$ \\
SmoothSA            & $44.8_{\pm1.6}$ & $43.4_{\pm0.7}$ \\
\textbf{TSA (ours)} & $\mathbf{45.6_{\pm3.8}}$ & $\mathbf{46.1_{\pm0.6}}$ \\
\bottomrule
\end{tabular}
\end{center}
\end{table}

\paragraph{Discussion.}
Tables~\ref{tab:ytvis_recognition} and~\ref{tab:ytvis_dynamics} show that TSA preserves discriminative object information while improving temporal predictability. For object recognition (Table~\ref{tab:ytvis_recognition}), TSA achieves the best classification performance, improving Top-1 accuracy to 91.3, compared with 90.5 for RandSF.Q, 90.4 for SmoothSA, and 85.8 for SlotContrast. It matches RandSF.Q on Top-3 accuracy (97.9 for both) and remains above SmoothSA (97.6) and clearly above SlotContrast (95.8). Since recognition is a per-frame appearance test, these results indicate that the activation mechanism does not trade away per-frame semantic content in exchange for temporal consistency. The bounding-box IoU of TSA (51.5) is on par with the best baseline (51.5 for SlotContrast) and higher than RandSF.Q (50.6) and SmoothSA (42.6). TSA matches fewer ground-truth instances than the baselines (7908 versus 8957, 8979, and 9249), reflecting that fewer slots participate in matching.

The advantage of TSA is more pronounced in object dynamics prediction (Table~\ref{tab:ytvis_dynamics}). TSA achieves 45.6 FG-ARI and 46.1 mBO, outperforming SmoothSA by $+0.8$ FG-ARI and $+2.7$ mBO, RandSF.Q by $+7.4$ FG-ARI and $+2.4$ mBO, and SlotContrast by $+16.1$ FG-ARI and $+12.9$ mBO. Since SlotFormer is trained on top of frozen slot representations, these gains reflect the quality of the learned slot trajectories rather than changes in the downstream predictor. The large improvement in rollout FG-ARI over RandSF.Q and SlotContrast indicates that TSA produces slots with more stable object correspondence across time, making future slot states easier to predict. Together, the recognition and dynamics results show that TSA retains strong per-frame object information while providing more temporally consistent representations for prediction, and that these representations transfer effectively to tasks beyond the primary object-centric segmentation setting.
\section{Additional qualitative results}
\label{app:qualitative}

\subsection{Additional comparisons with prior methods}
\label{app:compare}

\begin{figure}[t]
    \centering
    \includegraphics[width=\linewidth]{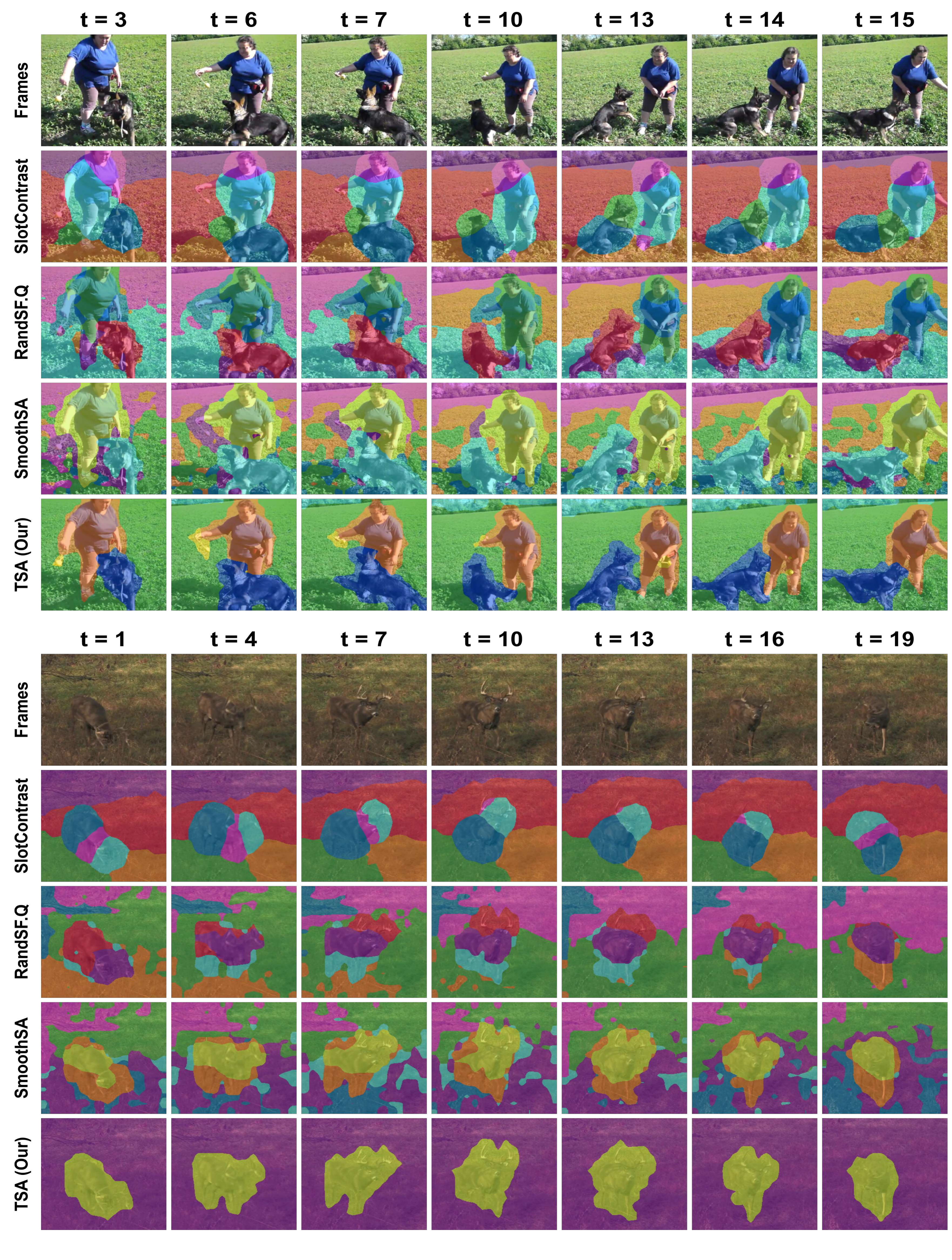}
    \caption{\textbf{Additional qualitative results on YouTube-VIS HQ.}}
    \label{fig:qual_ytvis_hq}
\end{figure}

\begin{figure}[t]
    \centering
    \includegraphics[width=\linewidth]{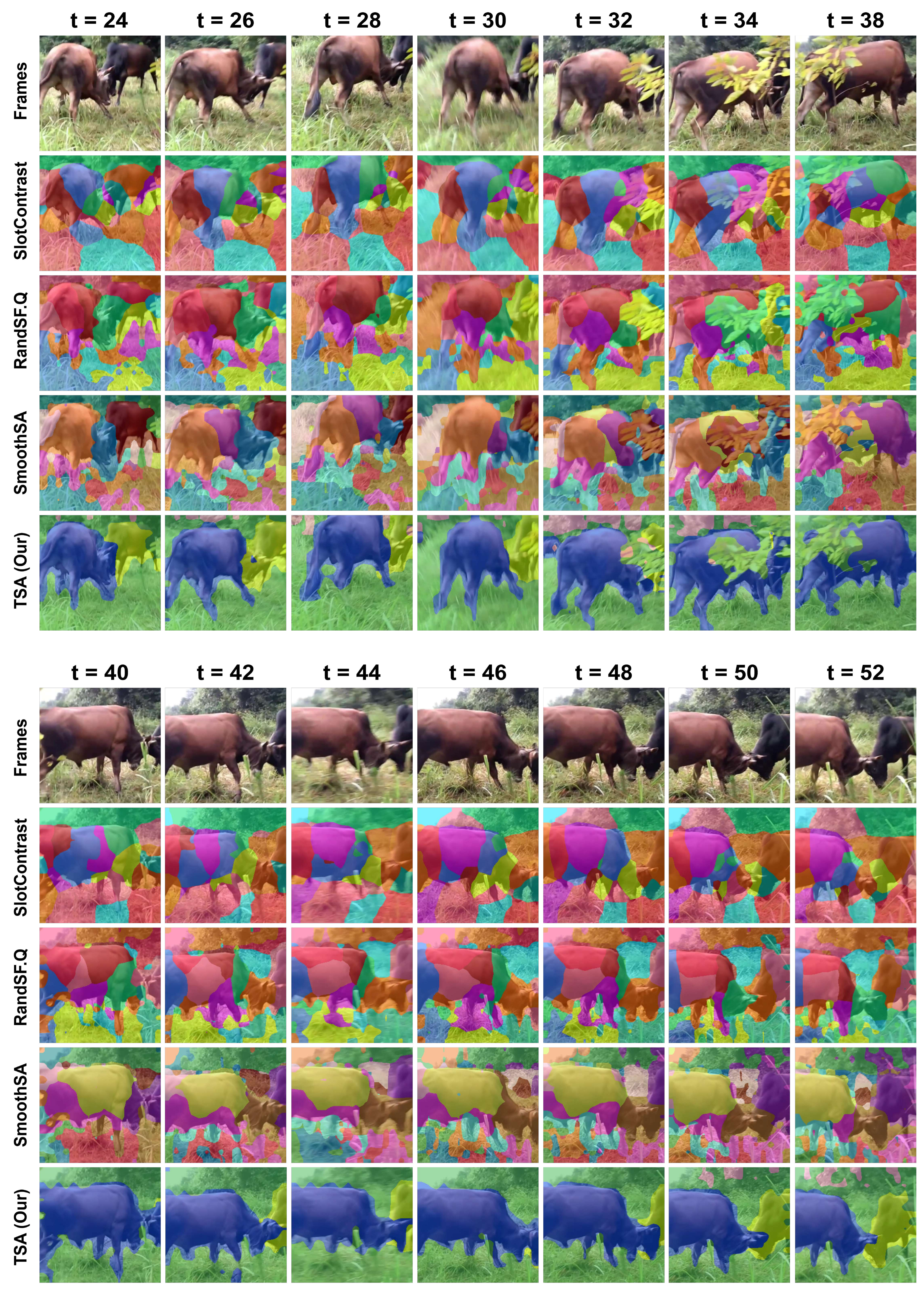}
    \caption{\textbf{Qualitative results on OVIS.}}
    \label{fig:qual_ovis}
\end{figure}

\begin{figure}[t]
    \centering
    \includegraphics[width=\linewidth]{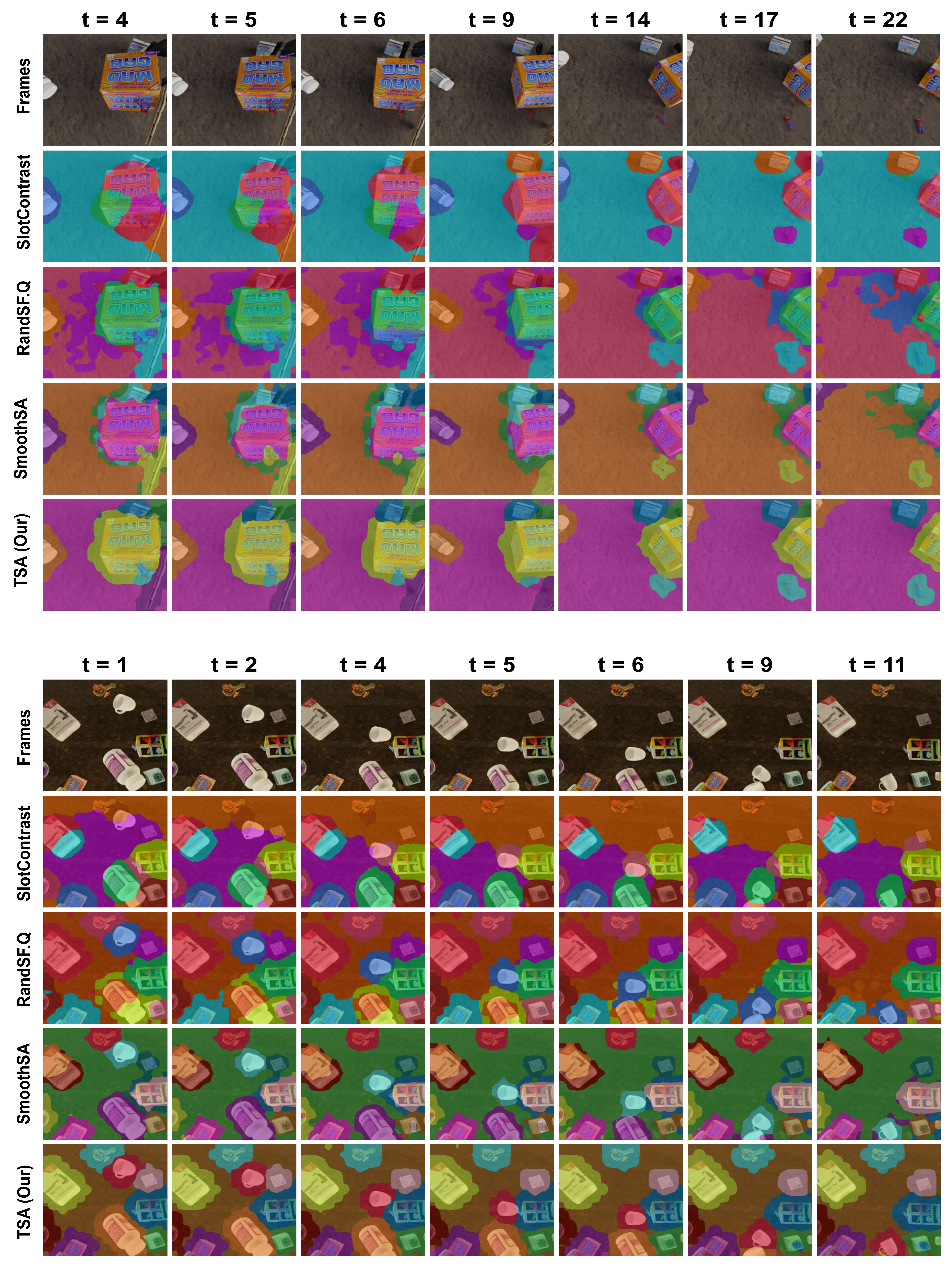}
    \caption{\textbf{Qualitative results on MOVi-C and MOVi-E.}}
    \label{fig:qual_movi}
\end{figure}

We compare TSA against RandSF.Q~\citep{zhao2026predicting},
SlotContrast~\citep{manasyan2025temporally}, and SmoothSA~\citep{zhao2026smoothsa} on YouTube-VIS HQ (Fig.~\ref{fig:qual_ytvis_hq}), OVIS (Fig.~\ref{fig:qual_ovis}), and MOVi-C/E (Fig.~\ref{fig:qual_movi}). Figs.~\ref{fig:qual_ovis} and~\ref{fig:qual_movi} each isolate one of the behaviors predicted in Sec.~\ref{sec:problem}, and Fig.~\ref{fig:qual_ytvis_hq} examines whether the same stability holds without occlusion.

\paragraph{Slot identity is preserved through occlusion and absence.}
In the two-animal sequence of Fig.~\ref{fig:qual_ovis} ($t{=}24$--$52$), the second animal is
first occluded by the leading animal, then hidden by foliage, and finally leaves the field of
view between $t{=}34$ and $t{=}40$. TSA maintains the two animals on two separate slots
throughout the sequence, and the same slot reacquires the animal upon its reappearance at
$t{=}42$. RandSF.Q, SlotContrast, and SmoothSA instead fragment each animal into several slots
whose assignment changes from frame to frame. The behavior of TSA is the one predicted by
Eq.~\ref{eq:state_gate}: when the current frame provides no evidence for the object, its slot
state is preserved rather than overwritten, and therefore remains a valid query at
reappearance.

\paragraph{A slot does not come to explain content other than its own object.}
Fig.~\ref{fig:qual_movi} (top, MOVi-C, $t{=}4$--$22$) illustrates the complementary behavior.
TSA holds the small object at the bottom of the scene on the same teal slot across the entire
sequence, and the thin rod on the same dark purple slot for as long as it remains visible,
whereas in the baselines the slots previously bound to these objects spread onto unrelated
regions. The same failure is visible at scene level on MOVi-E ($t{=}1$--$11$), where
SlotContrast assigns the background together with several objects to a single slot, while TSA
separates the individual objects and retains one background slot. Silencing a slot in the
decoder (Eq.~\ref{eq:log_bias}) is what prevents it from accounting for content it does not
represent.

\paragraph{The same stability is observed in the absence of occlusion.}
Fig.~\ref{fig:qual_ytvis_hq} shows two sequences in which the objects remain visible throughout.
In the sequence at the top ($t{=}3$--$15$), a person, a dog and a small yellow toy move in close
contact at very different scales: TSA assigns one slot to each and a single slot to the
background across all frames, whereas SlotContrast splits the person into several part-level
slots. RandSF.Q and SmoothSA, in turn, fail to separate the toy as its own object and leave the
background scattered with noisy slot assignments that change from frame to frame. The deer
sequence (bottom) shows the same contrast on a single object: SlotContrast, RandSF.Q, and
SmoothSA each split the deer into several slots, while TSA keeps it on one. Activation gating is
therefore not confined to occlusion-heavy settings: it also keeps objects whole and the scene
partition steady when every object stays in view.

\subsection{Ablation visualizations}
\label{app:qualitative_ablation}

\begin{figure}[t]
    \centering
    \includegraphics[width=\linewidth]{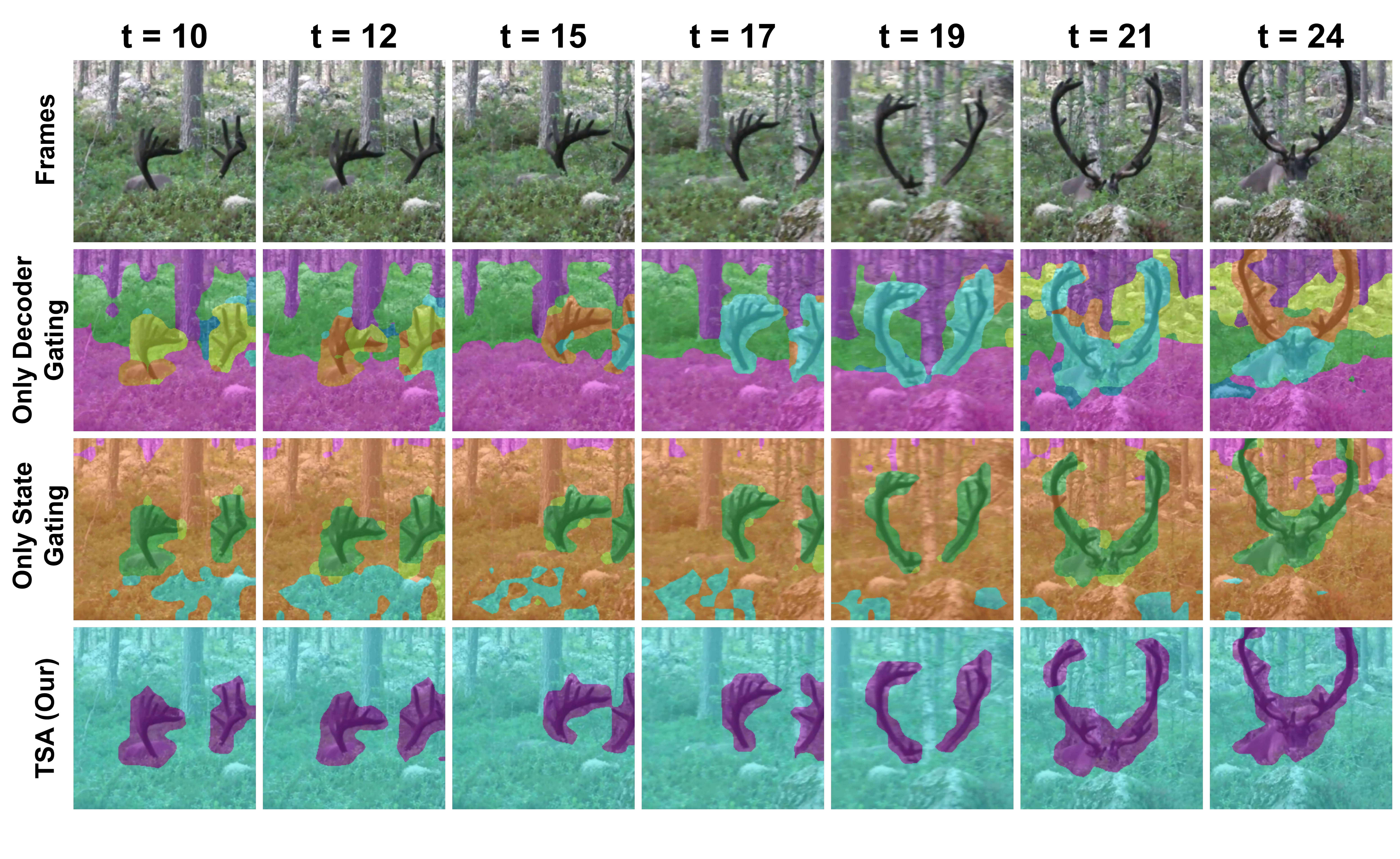}
    \caption{\textbf{Activation pathway ablations.} Comparison of TSA with activation-gated decoder participation only, activation-gated state update only, and both pathways jointly gated.}
    \label{fig:qual_ablation_coupling}
\end{figure}

\begin{figure}[t]
    \centering
    \includegraphics[width=\linewidth]{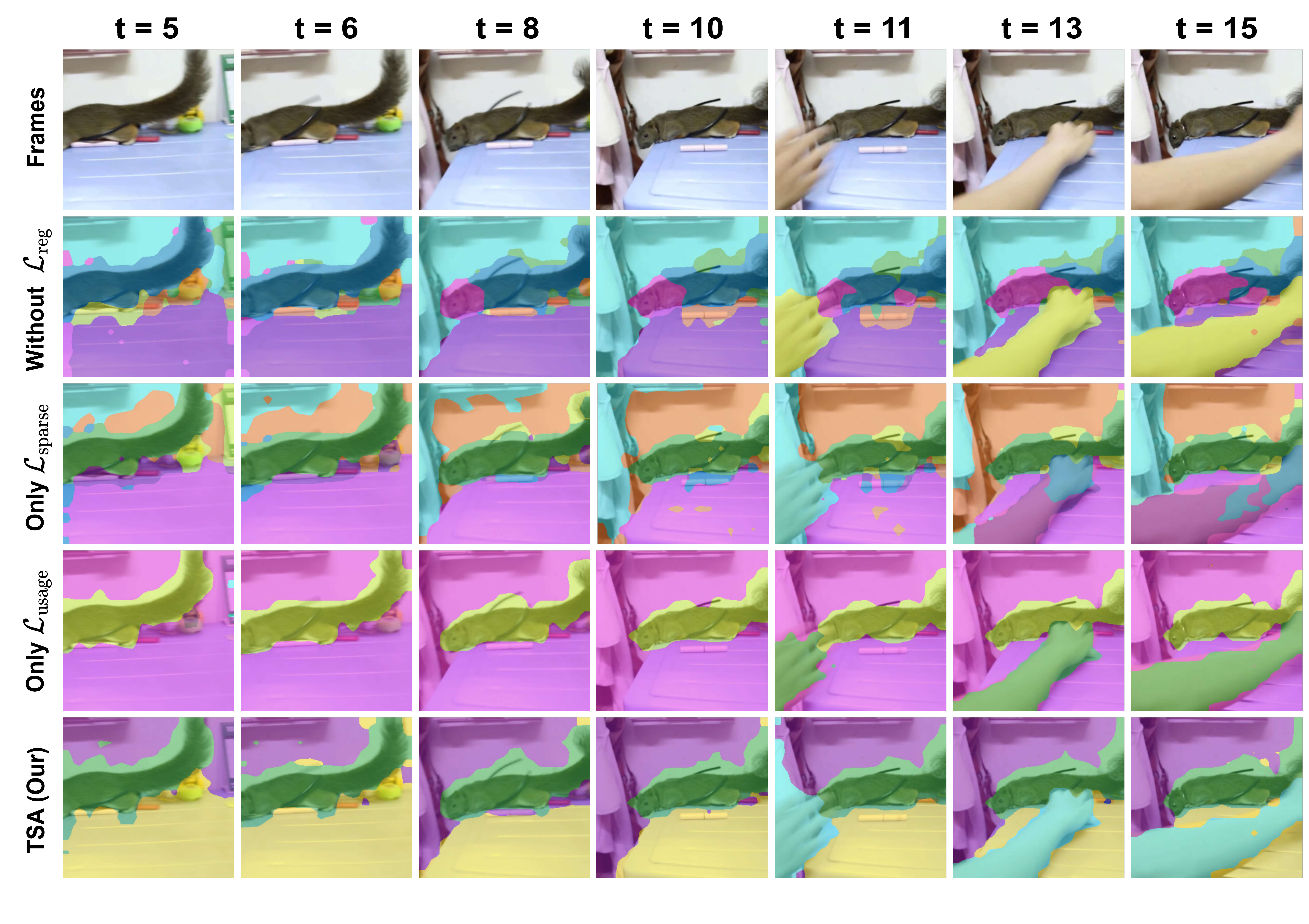}
    \caption{\textbf{Activation regularization ablations.} Comparison of TSA trained without $\Ls_{\mathrm{reg}}$, with $\Ls_{\mathrm{sparse}}$ only, with $\Ls_{\mathrm{usage}}$ only, and with both terms combined.}
    \label{fig:qual_ablation_regularization}
\end{figure}

\begin{figure}[t]
    \centering
    \includegraphics[width=\linewidth]{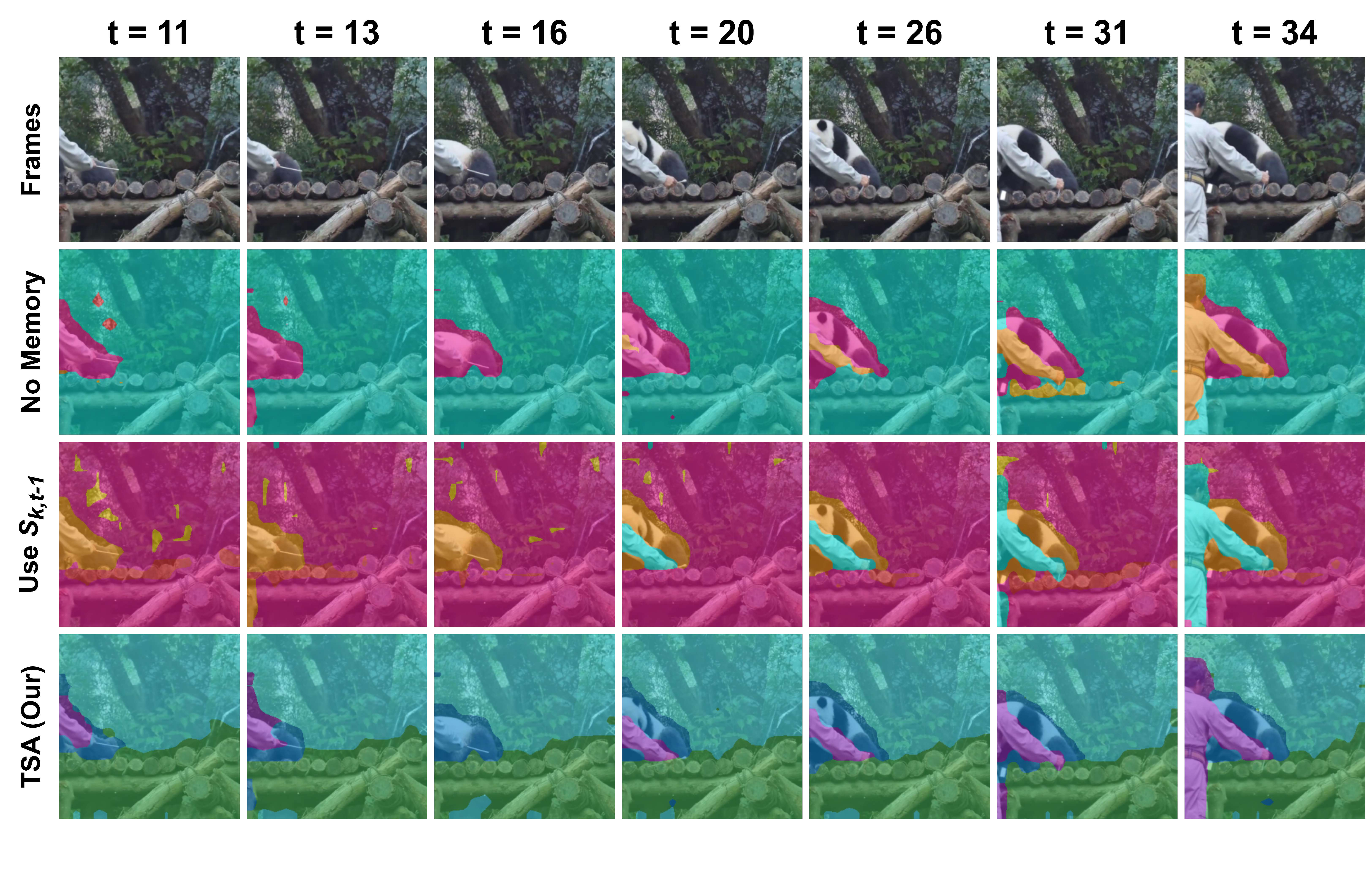}
    \caption{\textbf{Temporal memory ablation.} Comparison of different inputs to the Slot Activation Estimator $\Phi_{\mathrm{act}}$: no memory, the previous slot state $\vs_{k,t-1}$, and the temporal memory $\vm_{k,t-1}$ from the Temporal Context Encoder $\Psi_{\mathrm{tce}}$.}
    \label{fig:qual_ablation_memory}
\end{figure}

Fig.~\ref{fig:qual_ablation_coupling}--\ref{fig:qual_ablation_memory} provide qualitative evidence for the design choices studied quantitatively in Sec.~\ref{subsec:ablation}. All examples are drawn from YouTube-VIS HQ, using the ablation models of Appendix~\ref{app:ablation_ytvis}.

\paragraph{Effect of activation-gated state update and decoder participation.}
Fig.~\ref{fig:qual_ablation_coupling} visualizes the three gated configurations on a deer sequence. With decoder gating alone, the antlers are
split across several part-level slots and the background is repartitioned from frame to frame:
suppressing a slot in the decoder does not stop its state from being overwritten by
current-frame evidence, so each slot no longer stays on the same object. Gating the state update
alone keeps the antlers on a single slot across the whole sequence, confirming that anchoring
the state is what stabilizes identity, but the background is still scattered with spurious slot
activations. Gating both pathways holds the antlers on one slot and the background on one slot
throughout, giving the cleanest decomposition and supporting the design that state evolution and
reconstruction should be jointly controlled by a shared activation score.

\paragraph{Effect of regularization terms.}
Fig.~\ref{fig:qual_ablation_regularization} illustrates the complementary roles of $\Ls_{\mathrm{usage}}$ and $\Ls_{\mathrm{sparse}}$. Without either term, the scene is divided into many slots and the object is not held consistently across frames. With $\Ls_{\mathrm{usage}}$ alone, slot assignments become temporally consistent across frames; however, the pressure to reduce the active slot count can lead to over-compression, where a single slot absorbs multiple objects. With $\Ls_{\mathrm{sparse}}$ alone, activations are sharpened toward binary decisions but redundant slots remain active, so the model behaves similarly to unconditional propagation. Combining both losses balances these effects: $\Ls_{\mathrm{usage}}$ enforces compact slot usage with stable temporal correspondence, while $\Ls_{\mathrm{sparse}}$ ensures decisive activation transitions without collapsing distinct objects into the same slot.

\paragraph{Effect of temporal memory.}
Fig.~\ref{fig:qual_ablation_memory} compares different inputs to the Slot Activation Estimator $\Phi_{\mathrm{act}}$ on a sequence with persistent partial occlusion. Without temporal memory, activation predictions rely solely on the current Slot Attention candidate $\tilde{s}_{k,t}$, which carries
no summary of how the slot has behaved over time, so slots still drift onto other objects. Conditioning on the previous slot state $\vs_{k,t-1}$ provides a short-term temporal prior and stabilizes the overall scene partition, but a single past frame is not enough to keep the assignment consistent and the same drift remains. The full model uses the temporal memory $\vm_{k,t-1}$ from the Temporal Context Encoder $\Psi_{\mathrm{tce}}$, which summarizes the recent slot trajectory and keeps each slot on its own object throughout the sequence.

\end{document}